\documentclass[wcp]{jmlr}

\usepackage{bbm}
\usepackage{multirow}
\usepackage{xfrac}
\usepackage[T1]{fontenc}

\usepackage{enumitem}
\usepackage{epstopdf}

\usepackage{bm}
\usepackage{csquotes}
\usepackage{algorithm}
\usepackage{algorithmic}
\newcommand{\dale}{\hat{f}_{\mathtt{DALE}}}
\newcommand{\ale}{f_{\mathtt{ALE}}}
\newcommand{\alep}{\hat{f}_{\mathtt{ALE}}}
\newcommand{\Xc}{\mathcal{X}_c}
\newcommand{\xc}{\mathbf{x}_c}
\newcommand{\Xcb}{\mathcal{X}_c}
\newcommand{\Xb}{\mathcal{X}}

\newcommand{\xci}{\mathbf{x}^i_{\mathbf{c}}}
\newcommand{\xb}{\mathbf{x}}
\newcommand{\R}{\mathbb{R}}
\newcommand{\E}{\mathbb{E}}
\newcommand{\Jac}{\mathbf{J}}

\usepackage{microtype}

\usepackage{tikz}
\usetikzlibrary{matrix,positioning,arrows.meta,arrows,fit,backgrounds,decorations.pathreplacing}

\tikzset{ mymat/.style={ matrix of math nodes, text height=2.5ex, text
depth=0.75ex, text width=6.00ex, align=center, column
sep=-\pgflinewidth, nodes={minimum height=5.0ex} }, mymats/.style={
mymat, nodes={draw,fill=#1} }, mymat2/.style={ matrix of math nodes,
text height=1.0ex, text depth=0.0ex, minimum width=5ex, % text
width=7.00ex, align=center, column sep=-\pgflinewidth }, }

\usetikzlibrary{shapes.geometric, arrows, backgrounds, scopes}
\usepackage{pgfplots} \pgfplotsset{width=6.75cm, compat=newest}
\usepackage[utf8]{inputenc} \DeclareUnicodeCharacter{2212}{−}
\usepgfplotslibrary{groupplots,dateplot}
\usetikzlibrary{patterns,shapes.arrows}

\usepackage{longtable}% for long tables

\usepackage{booktabs}
\makeatletter
\let\Ginclude@graphics\@org@Ginclude@graphics 
\makeatother

\jmlrvolume{189}
\jmlryear{2022}
\jmlrworkshop{ACML 2022}

\title[DALE:~Differential Accumulated Local Effects]{DALE:~Differential Accumulated Local Effects for efficient and accurate global explanations}

\author{\Name{Vasilis Gkolemis} \Email{gkolemis@hua.gr, vgkolemis@athenarc.gr}\\
  \addr~Harokopio University of Athens, IMIS ATHENA Research Center
  \AND
  \Name{Theodore Dalamagas} \Email{dalamag@athenarc.gr}\\
  \addr~IMIS ATHENA Research Center
  \AND
  \Name{Christos Diou} \Email{cdiou@hua.gr}\\
  \addr~Harokopio University of Athens
}

\editors{Emtiyaz Khan and Mehmet G\"{o}nen}

\begin{document}

\maketitle

\begin{abstract}
Accumulated Local Effect (ALE) is a method for accurately estimating feature effects, overcoming fundamental failure modes of previously-existed methods, such as Partial Dependence Plots. However, \textit{ALE's approximation}, i.e.~the method for estimating ALE from the limited samples of the training set, faces two weaknesses. First, it does not scale well in cases where the input has high dimensionality, and, second, it is vulnerable to out-of-distribution (OOD) sampling when the training set is relatively small. In this paper, we propose a novel ALE approximation, called Differential Accumulated Local Effects (DALE), which can be used in cases where the ML model is differentiable and an auto-differentiable framework is accessible. Our proposal has significant computational advantages, making feature effect estimation applicable to high-dimensional Machine Learning scenarios with near-zero computational overhead. Furthermore, DALE does not create artificial points for calculating the feature effect, resolving misleading estimations due to OOD sampling. Finally, we formally prove that, under some hypotheses, DALE is an unbiased estimator of ALE and we present a method for quantifying the standard error of the explanation. Experiments using both synthetic and real datasets demonstrate the value of the proposed approach.
\end{abstract}

\section{Introduction}
\label{sec:1-introduction}

Machine Learning (ML) models have been adopted to high-stakes application domains, such as healthcare and finance. These fields require methods with the ability to explain their predictions, i.e., justify why a specific outcome has emerged. However, several types of accurate and highly non-linear models like Deep Neural Networks do not meet this requirement. Therefore, there is a growing need for explainability methods for interpreting such ``black-box'' models. Feature effect forms a fundamental category of global explainability methods (i.e. characterizing the model as a whole, not a particular input). The goal of the feature effect is to isolate the average impact of a single feature on the output. This class of methods is attractive due to the simplicity of the explanation that is easily understandable by a non-expert.

There are three popular feature effect methods: (i) Partial Dependence Plots (PDPlots) \citep{Friedman2001}, (ii) Marginal Plots (MPlots)~\citep{Apley2020} and (iii) Accumulated Local Effects (ALE)~\citep{Apley2020}. PDPlots and MPlots assume that input features are not correlated. When this does not hold, both methods lead to misestimation; PDPlots quantify the effect by marginalizing over out-of-distribution (OOD) synthetic instances, and MPlots yield aggregated effects on single features. Therefore, both methods perform well only in independent or low-correlated features. ALE is the only feature effect method that succeeds in staying on distribution and isolating feature effects in situations where input features are highly correlated.\footnote{In Section~\ref{sec:3-feature-effect}, we provide a thorough analysis for clarifying the differences between these three approaches.} However, in most cases, it is impossible to compute ALE through its definition since this would require (a) solving a high-dimensional integral, which is infeasible, and (b) evaluating the data generating distribution, which is usually unknown. Therefore,~\cite{Apley2020} proposed an estimating ALE with a Monte-Carlo approximation. This approximation faces two weaknesses. First, it becomes computationally inefficient in cases of datasets with numerous high-dimensional instances. Second, it is still vulnerable to OOD sampling in cases of wide bin sizes.

This paper proposes Differential Accumulated Local Effects (DALE), a novel approximation for ALE that resolves both weaknesses. DALE leverages auto-differentiation for computing the derivatives wrt each instance in a single pass. Therefore, it scales well in the case of high-dimensional inputs, large training sets and expensive black-box models. Furthermore, DALE estimates the feature effect using only the examples from the training set, securing that the estimation is not affected by OOD samples.
The contributions of this work are:
\begin{itemize}
\item We introduce DALE, a novel approximation to efficiently create ALE plots on differentiable black-box models. DALE is more efficient than the traditional ALE approximation, scales much better to high-dimensional datasets, and avoids OOD sampling.
\item We formally prove that DALE is an unbiased estimator of ALE and quantify the standard error of the approximation.
\item We show with synthetic and real datasets that DALE: (a) scales
  better than ALE, (b) provides a better approximation than ALE,
  especially in cases of wide bins. Code for reproducing all
  experiments is provided at \href{https://github.com/givasile/DALE}{https://github.com/givasile/DALE}.
\end{itemize}

\section{Related Work}
\label{sec:2-related}

Explainable AI (XAI) is a fast-evolving field with a growing interest. In recent years, the domain has matured by establishing its terminology and objectives~\citep{Hoffman2018}. Several surveys have been published~\citep{BarredoArrieta2020},~\citep{Adadi2018} classifying the different approaches and detecting future challenges on the field~\citep{Molnar2020}. There are several criteria for grouping XAI methods. A very popular distinction is between local and global ones. Local interpretability methods explain why a model made a specific prediction given a specific input. For example, local surrogates such as LIME~\citep{Ribeiro2016} train an explainable-by-design model in data points generated from a local area around the input under examination. SHAP values~\citep{Lundberg2017} measures the contribution of each attribute in a specific prediction, formulating a game-theoretical framework based on Shapley Values. Counterfactuals~\citep{Wachter2017} search for a data point as close as possible to the examined input that flips the prediction. Anchors~\citep{Ribeiro2018} provide a rule, i.e., a set of attribute values, that is enough to freeze the prediction, independently of the value of the rest of the attributes.

Global methods, which is the focus of this paper, explain the average model behavior. Global surrogates approximate the black-box model with a simple one, for example, decision trees~\citep{nanfack2021global}. Prototypes~\citep{Gurumoorthy2019} search for data points that are representative for each class and criticisms~\citep{Kim2016} for ambiguous data points that representative of the boarder between classes. Global feature importance methods characterize each input feature by assigning to it an importance score. Permutation feature importance~\citep{Fisher2019} measures the change in the prediction score of a model, after permuting the value of each feature. Often, apart from knowing that a feature is important, it also valuable to know the type of the effect on the output (positive/negative). Feature effect methods take a step further and quantify the type of a each feature attribute influences the output on average. There are three popular feature effect techniques Partial Dependence Plots~\citep{Friedman2001}, Marginal Plots and ALE~\citep{Apley2020}. Another class of global explanation techniques measures the interaction~\citep{Friedman2008} between features. Feature interaction quantifies to what extent the effect of two variables on the outcome is because of their combination.~\cite{Friedman2008} proposed a set of appropriate visualizations for such interactions. The generalization of feature effect and variable interactions is functional decomposition~\citep{Molnar2021}, that decomposes the black-box function into a set of simpler ones that may include more than two features.

\section{Background}
\label{sec:3-feature-effect}

This section introduces the reader to three popular feature effect methods; PDPlots, MPlots and ALE.

\paragraph*{Notation.} We use uppercase and calligraphic font \( \mathcal{X}\) for random variables (rv), plain lowercase \( x \) for scalar variables and bold \( \xb \) for vectors. Often, we partition the input vector \(\xb \in \R^D\) to the feature of interest \(x_s \in \R \) and the rest of the features \(\xc \in \R^{D-1}\), and for convenience we notate it \(\xb = (x_s, \xc)\). We clarify that \((x_s, \xc)\) corresponds to the vector \((x_1, \cdots, x_s, \cdots x_D)\). Equivalently, we notate the corresponding rv to \(\mathcal{X} = (\mathcal{X}_s, \mathcal{X}_c)\). The black-box function is \( f: \R^D \rightarrow \R \) and the feature effect of the \(s\)-th feature is \(f_{\mathtt{<method>}}(x_s)\), where \(\mathtt{<method>}\) is the name of the feature effect method.% \footnote{An extensive list of all symbols used in the paper is provided in the helping material.}

\paragraph{Feature Effect Methods.} PDPlots formulate the feature effect of the \(s\)-th attribute as an expectation over the marginal distribution \(\mathcal{X}_c\), i.e., \(f_{\mathtt{PDP}}(x_s) = \mathbb{\E}_{\mathcal{X}_c}[f(x_s,\mathcal{X}_c)]\). MPlots formulate it as an expectation over the conditional \(\mathcal{X}_c|\mathcal{X}_s\), i.e., \(f_{\mathtt{MP}}(x_s) = \E_{\mathcal{X}_c|\mathcal{X}_s = x_s}[f(x_s, \mathcal{X}_c)]\). ALE computes the global effect at \(x_s\) as an accumulation of the local effects. The local effect at point \(z\) is the expected change on the output over the conditional distribution \(\Xcb|\mathcal{X}_s=z\), i.e.  \( \E_{\Xcb|\mathcal{X}_s=z} \left [ \frac{\partial f(x_s, \mathcal{X}_c)}{\partial x_s} \right] \). The formula that defines ALE is presented below:

\begin{gather}
  \label{eq:ALE} f_{\mathtt{ALE}}(x_s) = c + \int_{-\infty}^{x_s} \mathbb{E}_{\Xcb|\mathcal{X}_s=z}\left[\frac{\partial f(z, \mathcal{X}_c)}{\partial z}\right] \partial z
\end{gather}
The constant \(c\) is used for centering the ALE plot. For illustrating the differences between the methods and the superiority of ALE, we provide a toy example. We select a bivariate black-box function \(f\) with correlated features; the first feature \( x_1 \) follows a uniform distribution \( x_1 \sim \mathcal{U}(0,1)\) and the second feature gets the value of \(x_1\) in a deterministic way, i.e., \( x_2 = x_1 \). The black-box function is the following piece-wise linear mapping:
\begin{equation} \label{eq:example-1-mapping} f(x_1, x_2) =
  \begin{cases} 1 - x_1 - x_2 & x_1 + x_2 \leq 1 \\ 0 & \text{otherwise}
  \end{cases}
\end{equation}
\noindent
Due to the piece-wise linear form, it is easy to isolate the effect of \(x_1\); Insider the region \(0 \leq x_1 \leq 0.5\), the effect is linear, i.e., \(-x_1\), and outside it is constant, i.e., the effect does not depend on \(x_1\). The closed-form solution for each method is presented below:~\footnote{Detailed derivations can be found in the helping material.}~\footnote{ Due to symmetry, for each method, the effect for \(x_2\) is the same with the effect of \(x_1\)}
\begin{equation}\label{eq:example-1-pdp} f_{\mathtt{PDP}}(x_1) = \mathbb{\E}_{\mathcal{X}_2} \left [f(x_1,\mathcal{X}_2) \right] = \frac{{(1-x_1)}^2}{2}, \: \forall x_1 \in [0,1]
\end{equation}
\begin{equation} \label{eq:example-1-mplots}
  \begin{split} f_{\mathtt{MP}}(x_1) &= \E_{\mathcal{X}_2|\mathcal{X}_1 = x_1} \left [ f(x_1, \mathcal{X}_2) \right] = \begin{cases} 1 - 2x_1 & x_1 \leq 0.5 \\ 0 &\text{otherwise}
    \end{cases}
  \end{split}
\end{equation}
\begin{align}\label{eq:example-1-ale}
  \begin{split} f_{\mathtt{ALE}}(x_1) &= c + \int_{z_0}^{x_1} \E_{\mathcal{X}_2|\mathcal{X}_1=z} \left [ \frac{\partial f(z, \mathcal{X}_2)}{\partial z} \right] \partial z =
     \begin{cases} c - x_1 & 0 \leq x_1 \leq 0.5\\ c - 0.5 & 0.5 \leq x_1 \leq 1
    \end{cases}
  \end{split}
\end{align}
The effect computed in Eqs.~\eqref{eq:example-1-pdp},~\eqref{eq:example-1-mplots} helps us understand that PDPlots and MPlots provide misleading results in cases of correlated features. PDPlots integrate over unrealistic instances due to the use of the marginal distribution \( p(\mathcal{X}_1) \). Therefore, they incorrectly result in a quadratic effect in the region \(x_1 \in [0, 1]\). MPlots resolve this issue using the conditional distribution \( \mathcal{X}_2|\mathcal{X}_1 \) but suffer from computing combined effects. In the linear subregion, the effect is overestimated as \( -2x_1 \) which is the combined effect of both \( x_1 \) and \( x_2 \). As Eq.~\eqref{eq:example-1-ale} shows, ALE resolves both issues and provides the correct effect.

In real scenarios, we cannot obtain a solution directly from Eq.~\eqref{eq:ALE}. Therefore,~\cite{Apley2020} proposed a solution by splitting the \(x_s\) axis into bins, computing the local effects inside each bin with a Monte Carlo approximation, and, finally, averaging the bin effects. As we discuss extensively in Sections~\ref{sec:4-2-computational} and~\ref{sec:4-3-robustness}, this approximation does not scale well to high-dimensional datasets and is vulnerable to OOD sampling.

\section{Differential Accumulated Local Effects (DALE)}

In this section, we present DALE.~First, we formulate the expression for the first and second-order DALE and, then, we explain its computational benefits and its robustness to OOD sampling. Finally, we quantify the standard error of the DALE estimation.

\subsection{Definition of DALE}
\label{sec:4-1-DALE} As briefly discussed in Section~\ref{sec:3-feature-effect}, in most cases it is infeasible to compute ALE in an analytical form. Therefore,~\cite{Apley2020} proposed the following approximation that is based on the instances of the training set:

\begin{align} \hat{f}_{\mathtt{ALE}}(x_s) &= \sum_{k=1}^{k_x} \frac{1}{|\mathcal{S}_k|} \sum_{i:\xb^i \in \mathcal{S}_k} [f(z_k,\xci) - f(z_{k-1}, \xci)]
  \label{eq:ALE_appr}
\end{align}
We denote as \( \xb^i \) the \(i\)-th example of the training set and as \(x_s^i\) its \(s\)-th feature. \(k_x\) is the index of the bin \(x_s\) belongs to, i.e., \(k_x: z_{k_x-1} \leq x_s < z_{k_x} \) and \(\mathcal{S}_k\) is the set of points that lie in the \(k\)-th bin, i.e.  \( \mathcal{S}_k = \{ \xb^i : z_{k-1} \leq x^i_s < z_{k} \} \).  For understanding Eq.~\eqref{eq:ALE_appr} better, we split it in three levels: Instance effect is the effect computed on the \(i\)-th example, i.e., \(\Delta f_i = f(z_k,\xci) - f(z_{k-1}, \xci)\), bin effect is the effect computed by the samples that are in the \(k\)-th bin, i.e.  \(\frac{1}{|\mathcal{S}_k|} \sum_{i:\xb^i \in \mathcal{S}_k} \Delta f_i \), and global effect is ALE approximation \(\hat{f}_{\mathtt{ALE}}(x_s)\). The approximation splits the axis into \( K \) equally-sized bins and computes each bin effect by averaging the instance effects of the samples that lie in each bin. The global effect is the accumulation of the bin effects. To make a connection with ALE definition (Eq.~\ref{eq:ALE}), the bin effect is an estimation of the accumulated local effects of the interval covered by the bin, i.e.  \(\int_{z_{k-1}}^{z_k} \E_{\Xcb|\mathcal{X}_s=z}\left[\frac{\partial f(z,\Xcb)}{\partial x_s}\right] \partial z \).
The approach of Eq.~(\ref{eq:ALE_appr}) has some weaknesses. Firstly, it is computationally demanding since it evaluates \(f\) for \(2 \cdot N \cdot D\) artificial samples, where \(N\) is the number of samples in the dataset and \(D\) is the number of features.  Secondly, it is vulnerable to OOD sampling when the bins length becomes large. This happens because the instance effects are estimated by generating artificial samples at the bin limits. Finally, the whole computation usable only for a predefined bin length; altering the bin size for assessing the feature effect at a different resolution, requires all computations to be repeated from scratch.

\subsubsection{First-order DALE.}
To address these drawbacks, we propose Differential Accumulated Feature Effect (DALE) that exploits the partial derivatives without altering the data points. The following formula describes the first-order DALE approximation:
\begin{equation} \dale(x_s) = \Delta x \sum_{k=1}^{k_x} \frac{1}{|\mathcal{S}_k|} \sum_{i:\xb^i \in \mathcal{S}_k} [f_s(\xb^i)] = \Delta x \sum_{k=1}^{k_x} \hat{\mu}_k
 \label{eq:DALE}
\end{equation}
where \(\Delta x\) is the bin length and \(f_s\) the partial derivative wrt \(x_s\), i.e.  \(f_s = \frac{\partial f}{\partial x_s}\). We use \(\hat{\mu}_k^s = \frac{1}{|\mathcal{S}_k|} \sum_{i:\xb^i \in \mathcal{S}_k} [f_s(\xb^i)]\) to indicate the estimated \(k\)-th bin effect. DALE uses only the dataset samples and doesn't perturb any feature, securing that we estimate the bin effect from on-distribution (observed) data points. In Eq.~(\ref{eq:DALE}), the estimation of the instance effect at each training sample is independent from the bin size. Unlike ALE approximation, the number of the bins (hyperparameter K) affects only the resolution of the plot and \textbf{not} the instance effects. Finally, DALE enables computing the local effects \( f_s(\xb^i) \) for \(s = \{1, \ldots, D \}\), \(i = \{1, \ldots, N \}\) once, and reusing them to create ALE plots of different resolutions.  Therefore, the user may experiment with feature effect plots at many different resolutions, with near-zero computational cost.

\subsubsection{Second-order DALE.}

~\cite{Apley2020} also provide a formula for approximating the combined effect of a pair of attributes \(x_l, x_m\):\footnote{For completeness, we provide the second-order ALE definition in the helping material.}

\begin{equation}
  \label{eq:ALE2approx} \hat{f}_{\mathtt{ALE}}(x_l, x_m) = \sum_{p=1}^{p_x} \sum_{q=1}^{q_x} \frac{1}{|\mathcal{S}_{p,q}|} \sum_{i:\xb^i \in \mathcal{S}_{p,q}} \Delta^2 f_i
\end{equation}
where \( \Delta^2 f_i = [f(z_p, z_q, \xc) - f(z_{p-1}, z_q, \xc)] - [f(z_p, z_{q-1}, \xc) - f(z_{p-1}, z_{q-1}, \xc)] \). As before, instead of evaluating the second-order derivative at the limits of the grid, we propose accessing the second-order derivatives on the data points. The following formula describes the second-order DALE approximation:

\begin{equation} \dale(x_l, x_m) = \Delta x_l \Delta x_m \sum_{p=1}^{p_x} \sum_{q=1}^{q_x} \frac{1}{|\mathcal{S}_{p,q}|} \sum_{i:\xb^i \in \mathcal{S}_{p,q}}f_{l,m}(\xb^i) = \Delta x_l \Delta x_m \sum_{p=1}^{p_x} \sum_{q=1}^{q_x} \hat{\mu}_{p,q}^s
  \label{eq:DALE-2}
\end{equation}
where \( f_{l,m}(\xb) \) is the second-order derivative evaluated at \(\xb^i\), i.e.  \( f_{l,m}(\xb) = \dfrac{\partial^2f(x)}{\partial x_l \partial x_m} \), and \(\Delta x_l\), \(\Delta x_m\) correspond to the bin step for features \(x_l\) and \(x_m\), respectively. As in the first-order description, we use \( \hat{\mu}_{p,q}^s = \frac{1}{|\mathcal{S}_{p,q}|} \sum_{i:\xb^i \in \mathcal{S}_{p,q}}f_{l,m}(\xb^i)\) to express the local effect at the bin \( (p, q) \). DALE second-order approximation has the same advantages over ALE as in the first-order case; it is faster, protects from OOD sampling and permits multi-resolution plots, with near-zero additional cost.

\subsection{Computational Benefit}
\label{sec:4-2-computational}

DALE approximation has significant computational advantages. For estimating the feature effect of all features, our approach processes the \(N\) data points of the training set. In contrast, ALE approximation generates and processes \(2 \cdot N \cdot D\), weighting by a factor of \(D\) the computational complexity and the memory requirements. Therefore, DALE scales nicely in problems with high dimensionality as is the case in most Deep Learning setups. Our approach is built on the computation of the Jacobian matrix,

\begin{equation} \Jac =
  \begin{bmatrix} \nabla_{\xb}f(\xb^1) \\ \vdots \\ \nabla_{\xb}f(\xb^N)
  \end{bmatrix} =
  \begin{bmatrix} f_1(\xb^1) & \dots & f_D(\xb^1)\\ \vdots & \ddots & \vdots \\ f_1(\xb^N) & \dots & f_D(\xb^N)
  \end{bmatrix}
\label{eq:jacobian}
\end{equation}

\noindent where, as before, \( f_s(\xb^i) \) is the partial derivative of the \(s\)-th feature evaluated at the \(i\)-th training point. Automatic differentiation enables the computation of the gradients wrt all features in a single pass. Computing the gradient vector for a training example \(\xb^i\) wrt all features \( \nabla_{\xb}f(\xb^i) = [f_1(\xb), \cdots, f_D(\xb)] \) is computationally equivalent to evaluating \(f(\xb^i)\). Based on this observation, computing the whole Jacobian matrix costs \(\mathcal{O}(N)\). In contrast, in ALE, the evaluation of \(f\) for \(2 \cdot N \cdot D\) times costs \(\mathcal{O}(N \cdot D)\). Our method, also, takes advantage of all existing automatic differentiation frameworks which are optimized for computing the gradients efficiently.\footnote{For example, the computation of that Jacobian can be done in a single command using TensorFlow \( \mathtt{tf.GradientTape.jacobian(predictions, X)} \) and PyTorch \( \mathtt{torch.autograd.functional.jacobian(f, X)} \)} In Algorithm~\ref{alg:dale}, we present DALE in an algorithmic form. The algorithm needs as input: (a) the black-box function \(f\), (b) the derivative of \(\nabla_{\mathbf{x}} f \) and (c) the dataset \( \mathbf{X} \).\footnote{Technically, having access to \(\nabla_{\mathbf{x}} f \) is not a prerequisite, since the partial derivative \(\frac{\partial f}{\partial x}\) can be approximated numerically, with finite differences. However, in this case, the computational advantages are canceled.} The parameter \( K \) defines the resolution of the DALE plot. The algorithm returns a matrix \(\mathbf{A}\), where the cell \(\mathbf{A}_{s,j}\) contains the effect of the \(j\)-th bin of the \(s\)-th feature, i.e., \(\dale^s(x) = \mathbf{A}_{s,k_x} \). Steps 3-5 iterate over each attribute, therefore these steps have complexity \(\mathcal{O}(N \cdot D)\). However these steps involve relatively cheap operations (allocation, averaging and aggregation) in comparison with the computation of the Jacobian matrix. Finally, with matrix \(\mathbf{A}\) computed, evaluating \(\dale(x)\) requires only locating the bin \(k_x\) that \( x \) belongs to. The same computational advantage also hold for the second-order DALE. In the second-order we need to compute the Hessian Matrix instead of the Jacobian (Step 1) and to allocate the points in a 2D grid instead of the sequence of intervals (Step 3).

\begin{algorithm}[h]
\caption{DALE approximation}
\label{alg:dale} \textbf{Input}: \( f, \nabla_{\mathbf{x}} f, \mathbf{X} \) \\ \textbf{Parameter}: \( K \) \\ \textbf{Output}: \(\mathbf{A}\)
\begin{algorithmic}[1]
  % [1] enables line numbers
  \STATE Compute the Jacobian \(\Jac\) of Eq. \eqref{eq:jacobian}
  \FOR{\(s = 1, \ldots , D\)}
  \STATE Allocate points \( \Rightarrow \mathcal{S}_k \forall k \)
  \STATE Estimate local effect \( \Rightarrow \hat{\mu}_k^s \forall k\) of Eq.~\eqref{eq:DALE}
  \STATE Aggregate \( \Rightarrow \mathbf{A}_{s,j} = \Delta x\sum_{k=1}^{j} \hat{\mu}_k^s, j = 1, \ldots, K \)
  \ENDFOR
  \STATE \textbf{return} \(\mathbf{A}\) \textbf{||} Note that \( \dale(x) = \mathbf{A}_{s,k_x} \)
\end{algorithmic}
\end{algorithm}

\subsection{Robustness to out-of-distribution sampling}
\label{sec:4-3-robustness} 
\begin{figure}[h] \centering \resizebox{.35\columnwidth}{!}{\input{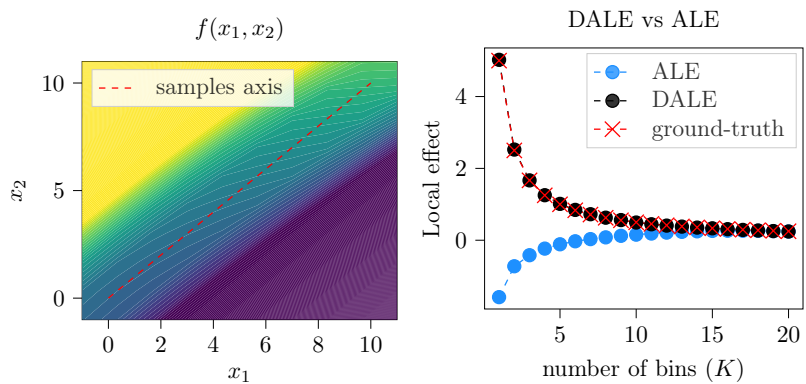}} \resizebox{.35\columnwidth}{!}{% This file was created with tikzplotlib v0.10.1.
\begin{tikzpicture}

\definecolor{darkgray176}{RGB}{176,176,176}
\definecolor{dodgerblue}{RGB}{30,144,255}
\definecolor{lightgray204}{RGB}{204,204,204}

\begin{axis}[
legend cell align={left},
legend style={fill opacity=0.8, draw opacity=1, text opacity=1, draw=lightgray204},
tick align=outside,
tick pos=left,
title={DALE vs ALE},
x grid style={darkgray176},
xlabel={number of bins \(\displaystyle (K)\)},
xmin=0.0499999999999999, xmax=20.95,
xtick style={color=black},
y grid style={darkgray176},
ylabel={Local effect},
ymin=-1.91455718867646, ymax=5.35298172476343,
ytick style={color=black}
]
\addplot [semithick, dodgerblue, dashed, mark=*, mark size=3, mark options={solid}]
table {%
1 -1.58421451079283
2 -0.724791391897207
3 -0.416931226166923
4 -0.233523181012335
5 -0.113002110067188
6 -0.0293908088400257
7 0.0312305603326059
8 0.0797401389995273
9 0.12275670716894
10 0.155053447821085
11 0.186088568383995
12 0.211864987771331
13 0.233108878251814
14 0.247869171569635
15 0.259566596177554
16 0.264108921400541
17 0.263569771086526
18 0.260892620884744
19 0.252751953555994
20 0.242460730712491
};
\addlegendentry{ALE}
\addplot [semithick, black, dashed, mark=*, mark size=3, mark options={solid}]
table {%
1 5.0226390468798
2 2.52393513932938
3 1.66605371806444
4 1.25321810135996
5 1.01182618748851
6 0.844347970008633
7 0.723984251300441
8 0.627612080163916
9 0.560862770386464
10 0.492398374084428
11 0.444051023514039
12 0.4070407257714
13 0.379038857924861
14 0.351678147881759
15 0.331857705495926
16 0.307844113455479
17 0.286910008925387
18 0.270908782811331
19 0.254748495290572
20 0.24246073071272
};
\addlegendentry{DALE}
\addplot [semithick, red, dashed, mark=x, mark size=5, mark options={solid}]
table {%
1 5
2 2.5
3 1.66666666666667
4 1.25
5 1
6 0.833333333333333
7 0.714285714285714
8 0.625
9 0.555555555555556
10 0.5
11 0.454545454545455
12 0.416666666666667
13 0.384615384615385
14 0.357142857142857
15 0.333333333333333
16 0.3125
17 0.294117647058824
18 0.277777777777778
19 0.263157894736842
20 0.25
};
\addlegendentry{ground-truth}
\end{axis}

\end{tikzpicture}}
  \caption[Example comparison]{(Left) The black-box function \(f\) of Section~\ref{sec:4-3-robustness}. (Right) Estimation of the bin effect of the first bin for DALE and ALE, for varying number of bins \(K\).}
  \label{fig:example-different-bins}
\end{figure}

OOD sampling is the source of failure in many explainability methods that perturb features(~\cite{Baniecki2022}, \cite{Hooker2021}). ALE is vulnerable to OOD sampling when the bin length is relatively big, or, equivalently, when the number of bins (hyperparameter \(K\)) is relatively small. We use the word \textit{relatively} to indicate that the threshold for characterizing a bin as big/small depends on the properties of the black-box function, i.e., how quickly it diverges outside of the data manifold. ML models learn to map \( \xb \rightarrow y \) only in the manifold of the data generating distribution \(\Xb\). Therefore, the black-box function \(f\) can take any arbitrary form away from \(\Xb\) without any increase in the training loss. On the other hand, when a limited number of samples is available, it maybe necessary to lower \(K\) to ensure a robust estimation of the mean effect. An end-to-end experimentation on the effect of OOD will be provided in Case 2 of Section~\ref{sec:5-1-artificial-experiments}.
In Figure~\Ref{fig:example-different-bins} we illustrate a small example where the underlying black-box function \(f\) has different behavior on the data generating distribution and away from it. As can be seen in Figure~\ref{fig:example-different-bins}(a), we set the black-box function to be \(f = x_1x_2\) inside \(|x_1-x_2| < 0.5\) and to rapidly diverge outside of this region. The first feature follows a uniform distribution, i.e., \(x_1 \sim U(0,10)\), and for the second feature \(x_2=x_1\). The local effect of \(x_1\) is \(\E_{x_2|x_1} \left [ f_1(\xb) \right ] = x_1 \). Splitting in \(K\) bins, the first bin covers the region \( [0, \frac{10}{K} ) \), therefore, as discussed in~\ref{sec:4-1-DALE}, the ground truth bin effect is \(\int_0^{10/K} \E_{x_2|z}\left[f_1(\xb)\right]\partial z = \frac{5}{K}\). In Figure~\Ref{fig:example-different-bins}(b), we observe that as the bin-length becomes bigger (smaller \(K\)), DALE approximates the effect perfectly, whereas, ALE fails due to OOD sampling. This happens because in the ALE approximation of Eq.~\eqref{eq:ALE_appr}, the bin limits \(z_{k-1}, z_k\) fall outside of the region \(|x_1-x_2| < 0.5\).

\subsection{Bias and variance}
\label{sec:4-4-std} Let a finite dataset of samples \(\mathcal{S}\), drawn independently and identically distributed (i.i.d) from the data generating distribution of \(\mathcal{X}\). DALE computes the accumulated local effect (Eq.~\eqref{eq:ALE}), using the approximation in (Eq.~\eqref{eq:DALE}). The expected value of the approximation across different datasets is

\begin{equation} \mathbb{E}_{\mathcal{S}}[\dale(x)] = \Delta x\sum_{k=1}^{k_x}\mathbb{E}_{\mathcal{S}}[\frac{1}{|\mathcal{S}_k|}\sum_{i:\xb^i \in \mathcal{S}_k} f_s(\xb^i)]
  \label{eq:bias_dale}
\end{equation}

\noindent Notice also that for the values of \(x\) at the end of bin \(k_x\), Eq.~\eqref{eq:ALE} can be rewritten as (after omitting the constant \(c\))
\begin{equation} f_{\mathtt{ALE}}(x) = \sum_{k = 1}^{k_x}\int_{x_{k-1}}^{x_k} \mathbb{E}_{\Xcb|\mathcal{X}_s=z}[f_s(\xb)] \partial z
    \label{eq:bias_ale_1}
\end{equation} where \(x_0=x_{s, min}\) and \(x_i\), \(i=1, \dotsc, k_x\) are the bin limits.

\noindent If we assume that each bin is sufficiently small such that \(f_s(\xb)\) does not depend on \(x_s\) (i.e., \(f(x)\) is linear wrt \(x_s\)) within the bin, then Eq. \eqref{eq:bias_ale_1} becomes
\begin{equation} f_{\mathtt{ALE}}(x) = \sum_{k = 1}^{k_x}\mathbb{E}_{\Xcb|\mathcal{X}_s \in \mathcal{S}_k}[f_s(\xb)]\int_{x_{k-1}}^{x_k} \partial z = \Delta x\sum_{k=1}^{k_x}\mathbb{E}_{\mathcal{X} \in \mathcal{S}_k}[f_s(\xb)]
    \label{eq:bias_ale_2}
\end{equation}

\noindent From Eqs. \eqref{eq:bias_dale} and \eqref{eq:bias_ale_2} we have
\begin{multline} \mathbb{E}_{\mathcal{S}}[\dale] - f_{\mathtt{ALE}}(x) = \Delta x\sum_{k=1}^{k_x}\mathbb{E}_{\mathcal{S}}[\frac{1}{|\mathcal{S}_k|}\sum_{k:\xb^k \in \mathcal{S}_k} f_s(\xb^k)] - \\ \Delta x\sum_{k=1}^{k_x}\mathbb{E}_{\mathcal{X}\in \mathcal{S}_k}[f_s(\xb)] = \Delta x\sum_{k=1}^{k_x}\left(\mathbb{E}_{\mathcal{S}}[\hat{\mu}_k^s] - \mu_k^s\right) = 0
\end{multline} since the expected value of the sample mean is an unbiased estimator of \(\mu_k^s\). As a result, under the condition of linearity wrt \(x_s\) within the bin, DALE is an unbiased estimator of the feature effect. If this assumption is violated (e.g., large bin size or highly nonlinear function), then this approach may introduce bias. The variance of the estimator is given\footnote{We show that in the supporting material.} by \( \mathrm{Var}[\hat{\mu}_k^s] = \dfrac{(\sigma_k^s)^2}{|\mathcal{S}_k|} \), where \((\sigma_k^s)^2\) is the variance of \(f_s\) within the bin. Furthermore, since the samples \(\xb^i\) are independent, \(\hat{\mu}_k^s\) for \(k=1,\dotsc,k_x\) are also independent. The variance of the estimation can then be approximated as
\begin{equation} \mathrm{Var}[\dale(x)] = (\Delta x)^2\sum_k^{k_x} \mathrm{Var} [\hat{\mu}_k^s] = (\Delta x)^2 \sum_k^{k_x} \dfrac{(\sigma_k^s)^2}{|\mathcal{S}_k|} \approx (\Delta x)^2 \sum_k^{k_x} \dfrac{(\hat{\sigma}_k^s)^2}{|\mathcal{S}_k|}
  \label{eq:DALE-var}
\end{equation}
where \((\hat{\sigma}_k^s)^2\) is the sample variance within bin \(k\). Equation~\eqref{eq:DALE-var} allows the calculation of the standard error for the DALE approximation.

% \subsection{Limitations of DALE}
% \label{sec:3-5-limitations} % \input{./chapters/3-5-limitations.tex}

\section{Experiments}

This section presents the experimental evaluation of DALE using two synthetic and one real dataset. The experiments aim to compare DALE (\(\dale\)) with ALE approximation (\(\hat{f}_{\mathtt{ALE}}\)) from the perspectives of both efficiency and accuracy.

\paragraph{Metrics.} For evaluating the efficiency of the approximations we measure the execution times (in seconds). For evaluating the accuracy we use: (a) qualitative comparison of the feature effect plots and (b) the Normalized Mean Squared Error which is defined as \(\text{NMSE}_{\mathtt{<approx>}} = \frac{\E[(\ale - f_{\mathtt{<approx>}})^2]}{\text{Var}[\ale]}\).

\paragraph{Synthetic Datasets.} The first synthetic dataset (Case 1) is designed to compare the approximations in terms of efficiency. For this reason, we generate design matrices \(\mathbf{X}\) of varying dimensionality \(D\) and number of instances \(N\). The second synthetic dataset (Case 2) is designed to compare the approximations in terms of accuracy. We define a data generating distribution \(\mathcal{X}\) and a black box function \(f\) with known forms for being able to directly compute the ground-truth ALE from Eq.~\eqref{eq:ALE}. Both \(\mathcal{X}\) and \(f\) are designed to amplify the effect of OOD sampling.

\paragraph{Real Dataset} We choose the Bike-Sharing dataset for two reasons. Firstly, it is the dataset utilized in the original ALE paper, so it is a proper choice for unbiased comparisons. Secondly, we wanted a dataset with enough training points to approximate the feature effect accurately, since the ground-truth is not available. Therefore, we want to check that \(\dale\) and \(\hat{f}_{\mathtt{ALE}}\) provide similar effects using dense bins. We also evaluate the accuracy of both methods behave when using larger bins.

\subsection{Synthetic Datasets}
\label{sec:5-1-artificial-experiments}
\subsubsection{Case 1 - Efficiency comparison}
\label{sec:case-1}

In this example, we evaluate the efficiency of the two approximations, \(\dale\) and \(\hat{f}_{\mathtt{ALE}}\), through the execution times. We want to compare how both approximations perform in terms of the dimensionality of the problem (\(D\)), the dataset size (\(N\)) and the size of the model \(L\). In each experiment we generate a design-matrix \( \mathbf{X} \), by drawing \( N \cdot D \) samples from a standard normal distribution. The black-box function \(f \) is a fully-connected neural network with \(L\) hidden layers of \( 1024 \) units each. All experiments are done using \(K=100\). We want to clarify that the value of \(K\) plays almost no role in the execution times.

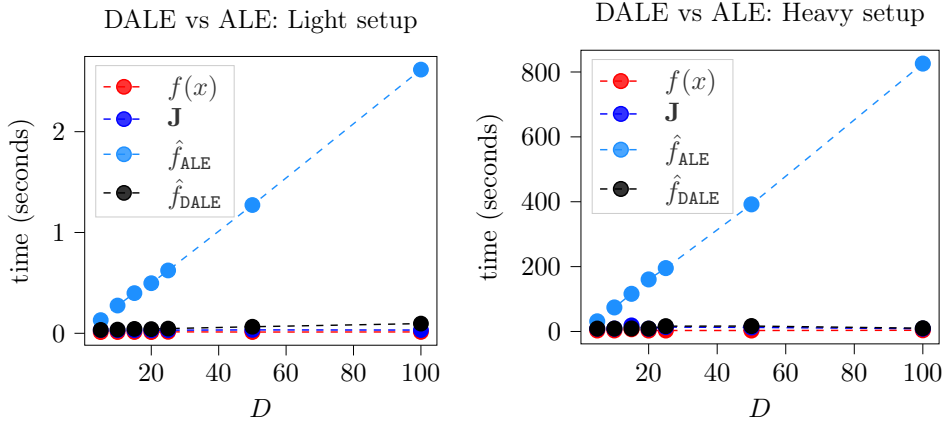
\begin{figure}[h]
  \centering
  \resizebox{.4\columnwidth}{!}{% This file was created with tikzplotlib v0.10.1.
\begin{tikzpicture}

\definecolor{darkgray176}{RGB}{176,176,176}
\definecolor{dodgerblue}{RGB}{30,144,255}
\definecolor{lightgray204}{RGB}{204,204,204}

\begin{axis}[
legend cell align={left},
legend style={
  fill opacity=0.8,
  draw opacity=1,
  text opacity=1,
  at={(0.03,0.97)},
  anchor=north west,
  draw=lightgray204
},
tick align=outside,
tick pos=left,
title={DALE vs ALE: Light setup},
x grid style={darkgray176},
xlabel={\(\displaystyle D\)},
xmin=0.25, xmax=104.75,
xtick style={color=black},
y grid style={darkgray176},
ylabel={time (seconds)},
ymin=-0.118335867500491, ymax=2.74596303149951,
ytick style={color=black}
]
\addplot [semithick, red, dashed, mark=*, mark size=3, mark options={solid}]
table {%
5 0.0118595361709595
10 0.0122641324996948
15 0.0128312110900879
20 0.0119729042053223
25 0.0144194364547729
50 0.0124425888061523
100 0.0135363340377808
};
\addlegendentry{$f(x)$}
\addplot [semithick, blue, dashed, mark=*, mark size=3, mark options={solid}]
table {%
5 0.0321429967880249
10 0.0295344591140747
15 0.0305428504943848
20 0.0298470258712769
25 0.0309284925460815
50 0.0336670875549316
100 0.032146692276001
};
\addlegendentry{$\mathbf{J}$}
\addplot [semithick, dodgerblue, dashed, mark=*, mark size=3, mark options={solid}]
table {%
5 0.130752086639404
10 0.275371313095093
15 0.398597836494446
20 0.496960043907166
25 0.623412370681763
50 1.27240407466888
100 2.61576771736145
};
\addlegendentry{$\hat{f}_{\mathtt{ALE}}$}
\addplot [semithick, black, dashed, mark=*, mark size=3, mark options={solid}]
table {%
5 0.0335890054702759
10 0.0366946458816528
15 0.0436010360717773
20 0.0414165258407593
25 0.0466822385787964
50 0.0650126934051514
100 0.0961912870407104
};
\addlegendentry{$\hat{f}_{\mathtt{DALE}}$}
\end{axis}

\end{tikzpicture}}
  \resizebox{.43\columnwidth}{!}{% This file was created with tikzplotlib v0.10.1.
\begin{tikzpicture}

\definecolor{darkgray176}{RGB}{176,176,176}
\definecolor{dodgerblue}{RGB}{30,144,255}
\definecolor{lightgray204}{RGB}{204,204,204}

\begin{axis}[
legend cell align={left},
legend style={
  fill opacity=0.8,
  draw opacity=1,
  text opacity=1,
  at={(0.03,0.97)},
  anchor=north west,
  draw=lightgray204
},
tick align=outside,
tick pos=left,
title={DALE vs ALE: Heavy setup},
x grid style={darkgray176},
xlabel={\(\displaystyle D\)},
xmin=0.25, xmax=104.75,
xtick style={color=black},
y grid style={darkgray176},
ylabel={time (seconds)},
ymin=-38.1810145603003, ymax=867.056094992299,
ytick style={color=black}
]
\addplot [semithick, red, dashed, mark=*, mark size=3, mark options={solid}]
table {%
5 3.13364505767822
10 3.24496150016785
15 6.86045789718628
20 2.96612668037415
25 3.05197858810425
50 3.02510833740234
100 3.71672296524048
};
\addlegendentry{$f(x)$}
\addplot [semithick, blue, dashed, mark=*, mark size=3, mark options={solid}]
table {%
5 8.88222980499268
10 8.82908821105957
15 18.4925575256348
20 8.51719951629639
25 13.5228071212769
50 12.4858655929565
100 8.59451198577881
};
\addlegendentry{$\mathbf{J}$}
\addplot [semithick, dodgerblue, dashed, mark=*, mark size=3, mark options={solid}]
table {%
5 31.310697555542
10 74.1438674926758
15 116.133689880371
20 160.465118408203
25 195.409912109375
50 391.950622558594
100 825.908935546875
};
\addlegendentry{$\hat{f}_{\mathtt{ALE}}$}
\addplot [semithick, black, dashed, mark=*, mark size=3, mark options={solid}]
table {%
5 8.83595371246338
10 9.03882694244385
15 9.21370220184326
20 8.9499044418335
25 16.3313121795654
50 16.7110271453857
100 10.0340776443481
};
\addlegendentry{$\hat{f}_{\mathtt{DALE}}$}
\end{axis}

\end{tikzpicture}}
  \caption[Case-1-fig-1]{Case 1. Comparison of the execution time of DALE and ALE wrt dimensionality in two setups: (Left) Light setup; \(N=10^3, L=2\).  (Right) Heavy setup; \(N=10^5, L=6\)}
  \label{fig:case-1-plots-1}
\end{figure}

In Figure~\Ref{fig:case-1-plots-1}, we directly compare \(\dale\) and \(\hat{f}_{\mathtt{ALE}}\) in two different setups: in Figure~\Ref{fig:case-1-plots-1}(Left), we use a light setup of \(N=10^3\) examples and a model of \(L=2\) layers, whereas in Figure~\Ref{fig:case-1-plots-1}(Right), a heavier setup with \(N=10^5\) and \(L=6\). We observe that in both cases, DALE executes in near-constant time independently of \(D\), while ALE scales linearly with wrt \(D\), confirming our claims of Section~\ref{sec:4-2-computational}. The difference in the execution time reaches significant levels from a relatively small dimensionality. In the heavy setup, ALE needs almost a minute for \(D=20\), three minutes for \(D=50\), and 15 minutes for \(D=100\). In all these cases, DALE executes in a few seconds. Another critical remark is that DALE's execution time is almost identical to the computation of the Jacobian \( \Jac \), which is benefited by automatic differentiation. Hence, we confirm that the overhead of performing steps 3-5 of Algorithm~\ref{alg:dale} is a small fraction of the total execution time. Another consequence of this remark is that we can test many different bin sizes with near-zero computational cost.

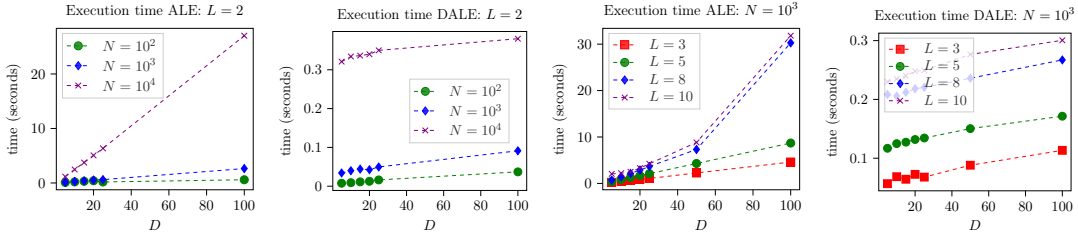
\begin{figure}[h]
  \centering
  \resizebox{.23\columnwidth}{!}{% This file was created with tikzplotlib v0.10.1.
\begin{tikzpicture}

\definecolor{darkgray176}{RGB}{176,176,176}
\definecolor{green01270}{RGB}{0,127,0}
\definecolor{lightgray204}{RGB}{204,204,204}
\definecolor{purple}{RGB}{128,0,128}

\begin{axis}[
legend cell align={left},
legend style={
  fill opacity=0.8,
  draw opacity=1,
  text opacity=1,
  at={(0.03,0.97)},
  anchor=north west,
  draw=lightgray204
},
tick align=outside,
tick pos=left,
title={Execution time ALE: \(\displaystyle L=2\)},
x grid style={darkgray176},
xlabel={\(\displaystyle D\)},
xmin=0.25, xmax=104.75,
xtick style={color=black},
y grid style={darkgray176},
ylabel={time (seconds)},
ymin=-1.24482673455045, ymax=28.3690538155494,
ytick style={color=black}
]
\addplot [semithick, green01270, dashed, mark=*, mark size=3, mark options={solid}]
table {%
5 0.101258754730225
10 0.202309727668762
15 0.302774310112
20 0.410390853881836
25 0.183210730552673
100 0.592442154884338
};
\addlegendentry{$N=10^2$}
\addplot [semithick, blue, dashed, mark=diamond*, mark size=3, mark options={solid}]
table {%
5 0.122078418731689
10 0.273244619369507
15 0.406428098678589
20 0.527456998825073
25 0.613609552383423
100 2.64824438095093
};
\addlegendentry{$N=10^3$}
\addplot [semithick, purple, dashed, mark=x, mark size=3, mark options={solid}]
table {%
5 1.15858960151672
10 2.47470235824585
15 3.75058650970459
20 5.05907726287842
25 6.34863948822021
100 27.0229682922363
};
\addlegendentry{$N=10^4$}
\end{axis}

\end{tikzpicture}}
  \resizebox{.23\columnwidth}{!}{% This file was created with tikzplotlib v0.10.1.
\begin{tikzpicture}

\definecolor{darkgray176}{RGB}{176,176,176}
\definecolor{green01270}{RGB}{0,127,0}
\definecolor{lightgray204}{RGB}{204,204,204}
\definecolor{purple}{RGB}{128,0,128}

\begin{axis}[
legend cell align={left},
legend style={
  fill opacity=0.8,
  draw opacity=1,
  text opacity=1,
  at={(0.91,0.5)},
  anchor=east,
  draw=lightgray204
},
tick align=outside,
tick pos=left,
title={Execution time DALE: \(\displaystyle L=2\)},
x grid style={darkgray176},
xlabel={\(\displaystyle D\)},
xmin=0.25, xmax=104.75,
xtick style={color=black},
y grid style={darkgray176},
ylabel={time (seconds)},
ymin=-0.0109329749522149, ymax=0.398615855950105,
ytick style={color=black}
]
\addplot [semithick, green01270, dashed, mark=*, mark size=3, mark options={solid}]
table {%
5 0.0076829195022583
10 0.00905656814575195
15 0.0112016201019287
20 0.0126841068267822
25 0.0161861181259155
100 0.0369504690170288
};
\addlegendentry{$N=10^2$}
\addplot [semithick, blue, dashed, mark=diamond*, mark size=3, mark options={solid}]
table {%
5 0.0340769290924072
10 0.0398837327957153
15 0.0440537929534912
20 0.0425540208816528
25 0.0495928525924683
100 0.0912438631057739
};
\addlegendentry{$N=10^3$}
\addplot [semithick, purple, dashed, mark=x, mark size=3, mark options={solid}]
table {%
5 0.320958256721497
10 0.334092974662781
15 0.33648693561554
20 0.340000033378601
25 0.350000023841858
100 0.379999995231628
};
\addlegendentry{$N=10^4$}
\end{axis}

\end{tikzpicture}}
  \resizebox{.23\columnwidth}{!}{% This file was created with tikzplotlib v0.10.1.
\begin{tikzpicture}

\definecolor{darkgray176}{RGB}{176,176,176}
\definecolor{green01270}{RGB}{0,127,0}
\definecolor{lightgray204}{RGB}{204,204,204}
\definecolor{purple}{RGB}{128,0,128}

\begin{axis}[
legend cell align={left},
legend style={
  fill opacity=0.8,
  draw opacity=1,
  text opacity=1,
  at={(0.03,0.97)},
  anchor=north west,
  draw=lightgray204
},
tick align=outside,
tick pos=left,
title={Execution time ALE: \(\displaystyle N=10^3\)},
x grid style={darkgray176},
xlabel={\(\displaystyle D\)},
xmin=0.25, xmax=104.75,
xtick style={color=black},
y grid style={darkgray176},
ylabel={time (seconds)},
ymin=-1.36616255235276, ymax=33.4035613893502,
ytick style={color=black}
]
\addplot [semithick, red, dashed, mark=square*, mark size=3, mark options={solid}]
table {%
5 0.214279413223267
10 0.445428252220154
15 0.637093186378479
20 0.876859426498413
25 1.06718361377716
50 2.25691056251526
100 4.60215711593628
};
\addlegendentry{$L=3$}
\addplot [semithick, green01270, dashed, mark=*, mark size=3, mark options={solid}]
table {%
5 0.408349752426147
10 0.824460029602051
15 1.26256990432739
20 1.68054091930389
25 2.10548663139343
50 4.29162120819092
100 8.69534683227539
};
\addlegendentry{$L=5$}
\addplot [semithick, blue, dashed, mark=diamond*, mark size=3, mark options={solid}]
table {%
5 0.71037769317627
10 1.42690515518188
15 2.14693737030029
20 2.87370538711548
25 3.65433740615845
50 7.31730794906616
100 30.2758731842041
};
\addlegendentry{$L=8$}
\addplot [semithick, purple, dashed, mark=x, mark size=3, mark options={solid}]
table {%
5 2.06213808059692
10 2.20000004768372
15 2.5
20 3.30623149871826
25 4.2340202331543
50 8.82898712158203
100 31.8231201171875
};
\addlegendentry{$L=10$}
\end{axis}

\end{tikzpicture}}
  \resizebox{.23\columnwidth}{!}{% This file was created with tikzplotlib v0.10.1.
\begin{tikzpicture}

\definecolor{darkgray176}{RGB}{176,176,176}
\definecolor{green01270}{RGB}{0,127,0}
\definecolor{lightgray204}{RGB}{204,204,204}
\definecolor{purple}{RGB}{128,0,128}

\begin{axis}[
legend cell align={left},
legend style={
  fill opacity=0.8,
  draw opacity=1,
  text opacity=1,
  at={(0.03,0.97)},
  anchor=north west,
  draw=lightgray204
},
tick align=outside,
tick pos=left,
title={Execution time DALE: \(\displaystyle N=10^3\)},
x grid style={darkgray176},
xlabel={\(\displaystyle D\)},
xmin=0.25, xmax=104.75,
xtick style={color=black},
y grid style={darkgray176},
ylabel={time (seconds)},
ymin=0.0445653210494129, ymax=0.312711131950527,
ytick style={color=black}
]
\addplot [semithick, red, dashed, mark=square*, mark size=3, mark options={solid}]
table {%
5 0.0567537546157837
10 0.0685902833938599
15 0.0642098188400269
20 0.0725626945495605
25 0.0678497552871704
50 0.088062047958374
100 0.113367915153503
};
\addlegendentry{$L=3$}
\addplot [semithick, green01270, dashed, mark=*, mark size=3, mark options={solid}]
table {%
5 0.116896510124207
10 0.124809384346008
15 0.127277135848999
20 0.131840586662292
25 0.134345889091492
50 0.150314569473267
100 0.171453237533569
};
\addlegendentry{$L=5$}
\addplot [semithick, blue, dashed, mark=diamond*, mark size=3, mark options={solid}]
table {%
5 0.208409428596497
10 0.205777525901794
15 0.212119460105896
20 0.218204021453857
25 0.220425486564636
50 0.236039161682129
100 0.267152309417725
};
\addlegendentry{$L=8$}
\addplot [semithick, purple, dashed, mark=x, mark size=3, mark options={solid}]
table {%
5 0.230000019073486
10 0.235000014305115
15 0.240000009536743
20 0.247859835624695
25 0.249285697937012
50 0.276461362838745
100 0.300522685050964
};
\addlegendentry{$L=10$}
\end{axis}

\end{tikzpicture}}
  \caption[Case-1-fig-2]{
    Case 1. Measurements of the execution time wrt dimensionality \(D\). From left to right:
    (a) \(\hat{f}_{\mathtt{ALE}}\) for \(L = 2\) and many dataset sizes \(N\)
    (b) \(\dale\) for \(L = 2\) and many dataset sizes \(N\)
    (c) \(\hat{f}_{\mathtt{ALE}}\) for \(N = 10^3\) and many model sizes \(L\)
    (d) \(\dale\) for \(N = 10^3\) and many model sizes \(L\)
  }
  \label{fig:case-1-plots-2}
\end{figure}

In Figure~\Ref{fig:case-1-plots-2}, we rigorously quantify to what extent the dataset size \(N\) and the model size \(L\) affect both methods. In Figures~\Ref{fig:case-1-plots-2}(a) and ~\Ref{fig:case-1-plots-2}(c), we confirm that both \(N\) and \(L\) have crucial impact in ALE's execution times. Therefore, for a big dataset and a heavy model \(f\), ALE's execution time quickly reaches prohibitive levels. In contrast, in Figures~\Ref{fig:case-1-plots-2}(b) and Figures~\Ref{fig:case-1-plots-2}(d), DALE is negligibly affected by these parameters. In the figures, we restrict the experiment to cases up to 100-dimensional input for illustration purposes. The same trend continues for an arbitrary number of dimensions. DALE can scale efficiently to any dimensionality as long as we have enough resources to store the dataset, evaluate the prediction model \(f\) and apply the gradients \(\nabla_{\xb}f\).

\subsubsection{Case 2 - Accuracy Comparison}
\label{sec:example2}

In this example, we evaluate the accuracy of the two approximations, \(\dale\) and \(\hat{f}_{\mathtt{ALE}}\), in a synthetic dataset where the ground truth ALE is accessible. As discussed in Section~\ref{sec:4-3-robustness}, ALE approximation is vulnerable to OOD sampling when the bins are wide, or equivalently, the number of bins \(K\) is small. We want to compare how both approximations behave in a case where the local effect is noisy.
We design an experiment where we know the black-box function and the data generating distribution. The black-box function \(f:\R^3 \rightarrow \R\) is split in three parts to amplify the effect of OOD sampling. It includes a mild term \( f_0(x) = x_1x_2 + x_1x_3 \) in the region \( 0 \leq |x_1 - x_2| < \tau \) and then a quadratic term \(g(x) = \alpha ((x_1 - x_2)^2 - \tau^2)\) is added(subtracted) over(under) the region, i.e.:

\begin{equation} \label{eq:example-2-mapping} f(\mathbf{x}) =
  \begin{cases} f_0(x) & , 0 \leq |x_1 - x_2| < \tau \\ f_0(x) - g(x) & , \tau \leq |x_1 - x_2| \\ f_0(x) + g(x) & , \tau \leq - |x_1 - x_2| \\
  \end{cases}
\end{equation}

\begin{figure}[h]
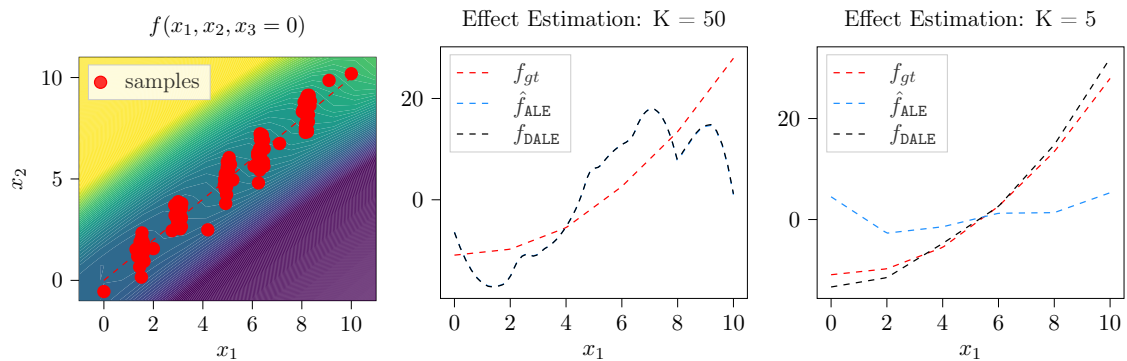

  \begin{center} \resizebox{.33\columnwidth}{!}{\input{./images/case-2-f-gt.tex}} \resizebox{.32\columnwidth}{!}{\input{./images/case-2-fe-bins-50.tex}} \resizebox{.32\columnwidth}{!}{\input{./images/case-2-fe-bins-5.tex}}
  \end{center}
  \caption[Case 2]{Case 2 experiment. (a) The black-box function \(f\) of Section~\ref{sec:4-3-robustness}. (b) Estimation of the feature effect for a large number of bins \(K=50\). (c) Estimation of the feature effect for a small number of bins \(K=5\).}
  \label{fig:example-2-samples}
\end{figure}

\noindent
The data points \(X^i = (x_1^i, x_2^i, x_3^i)\) are generated as follows; \(x_1^i \) are clustered around the points \(\{1.5, 3, 5, 7, 8.5\}\), \(x_2^i \sim \mathcal{N}(\mu=x_1, \sigma_2=0.1) \) and \(x_3^i \sim \mathcal{N}(\mu=0, \sigma_3^2=10) \). In Figure~\ref{fig:example-2-samples}(a), we illustrate \(f(\xb)\) for \(x_3=0\), as well as the generated data points. In this example, the local effect of \(x_1\) is \(\frac{\partial f}{\partial x_1} = x_2 + x_3\). Due to the noisy nature of \(x_3\), both ALE and DALE need a large number of sample for robust estimation. Therefore, we need to lower the number of bins \(K\). As will be shown below, both ALE and DALE fail to approximate the feature effect for high \(K\). On the other hand, when using a lower \(K\), ALE approximation fails due to OOD sampling, while DALE manages to accurately approximate the feature effect.

In Figure~\ref{fig:example-2-samples}(b) and Figure~\ref{fig:example-2-samples}(c), we observe the estimated effects for \(K=50\) and \(K=5\). In Figure~\ref{fig:example-2-samples}(b), \((K=50)\) the approximations converge to the same estimated effect which is inaccurate due to many noisy artifacts. In Figure~\ref{fig:example-2-samples}(c), \((K=5)\) we observe that for small \(K\), DALE approximates the ground-truth effect well, whereas ALE fails due to OOD sampling. Table~\ref{tab:case-2-accuracy} provides the NMSE of both approximation for varying number of bins \(K\). We observe that DALE consistently provides accurate estimations (NMSE \(\leq 0.1\)) for all small \(K\) values.

The experiments helps us confirm that when \(K \) increases, both approximations are based on a limited number of samples, and are vulnerable to noise. When \(K\) decreases, DALE lowers the resolution but provides more robust estimations. In contrast, ALE is vulnerable to OOD sampling.

\begin{table} \centering
  \caption{Case 2. Evaluation of the NMSE between the approximations and the ground truth. Blue color indicates the values that are below \(0.1\).}
  \label{tab:case-2-accuracy}
  \begin{tabular}{c|c|c|c|c|c|c|c|c|c}
    \multicolumn{10}{c}{Accuracy on the Synthetic Dataset (Case 2)} \\
    \hline \hline & & \multicolumn{8}{|c}{Number of bins} \\
    \hline & & 1 & 2 & 3 & 4 & 5 & 10 & 20 & 40 \\
    \hline \hline \multirow{2}{*}{\(\mathtt{NMSE}\)} & \(\alep\) & 100.42 & 22.09 & 4.97 & 2.81 & 0.78 & 1.49 & 0.34 & 0.39 \\
    & \(\dale\) & \textcolor{blue}{0.10} & \textcolor{blue}{0.03} & \textcolor{blue}{0.09} & \textcolor{blue}{0.02} & \textcolor{blue}{0.02} & 0.82 & 0.24 & 0.38 \\
    \hline
  \end{tabular}
\end{table}

\subsection{Real dataset}
\label{sec:5-2-real-datasets} In this section, we test our approximation on the Bike-Sharing Dataset~\cite{BikeSharing}.~\footnote{It is dataset drawn from the Capital Bikeshare system (Washington D.C., USA) over the period 2011-2012. The dataset can be found \href{https://archive.ics.uci.edu/ml/machine-learning-databases/00275/Bike-Sharing-Dataset.zip}{here}} The Bike-Sharing Dataset is chosen as the main illustration example in the original ALE paper, therefore it was considered appropriate for comparisons. The dataset contains the bike rentals for almost every hour over the period 2011 and 2012. The dataset contains 14 features, which we denote as \( X_{\mathtt{<feature\_name>}} \). We select the 11 features that are relevant to the prediction task. Most of the features are measurements of the environmental conditions, e.g.  \(X_{\mathtt{month}}\), \(X_{\mathtt{hour}}\), \(X_{\mathtt{temperature}}\), \(X_{\mathtt{humidity}}\), \(X_{\mathtt{windspeed}}\), while some others inform us about the day-type, e.g. whether we refer to a working-day \(X_{\mathtt{workingday}}\). The target value \( Y_{\mathtt{count}}\) is the bike rentals per hour, which has mean value \(\mu_{\mathtt{count}} = 189\) and standard deviation \(\sigma_{\mathtt{count}} = 181\). We train a deep fully-connected Neural Network with 6 hidden layers and \(711681\) parameters. We train the model for \(20\) epochs, using the Adam optimizer with learning rate 0.01. The model achieves a mean absolute error on the test of about \(38\) counts.

\paragraph{Efficiency.} For comparing the efficiency, we measure the execution time of DALE and ALE for a variable number of features. We present the results in Table~\ref{tab:bike-sharing-efficiency}. We confirm that DALE can compute the feature effect for all features in almost constant time wrt \(D\). In contrast, ALE scales linearly wrt \(D\) which leads to an execution time of over \(10\) seconds.

\begin{table}
  \caption{Bike-Sharing Dataset. Measurements of the execution time.}
  \label{tab:bike-sharing-efficiency} \centering
  \begin{tabular}{c|c|c|c|c|c|c|c|c|c|c|c}
    \multicolumn{12}{c}{Efficiency on Bike-Sharing Dataset (Execution Times in seconds)} \\
    \hline\hline & \multicolumn{11}{|c}{Number of Features} \\
    \hline & 1 & 2 & 3 & 4 & 5 & 6 & 7 & 8 & 9 & 10 & 11 \\
    \hline \( \alep \) & \textbf{0.85} & 1.78 & 2.69 & 3.66 & 4.64 & 5.64 & 6.85 & 7.73 & 8.86 & 9.9 & 10.9 \\
    \hline \( \dale \) & 1.17 & \textbf{1.19} & \textbf{1.22} & \textbf{1.24} & \textbf{1.27} & \textbf{1.30} & \textbf{1.36} & \textbf{1.32} & \textbf{1.33} & \textbf{1.37} & \textbf{1.39} \\
    \hline
  \end{tabular}
\end{table}

\paragraph{Accuracy.}

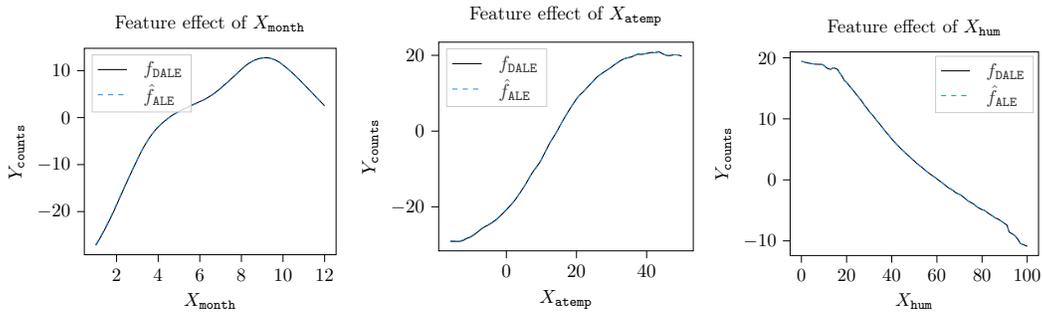
\begin{figure}[h]
  \centering
  \resizebox{.3\columnwidth}{!}{% This file was created with tikzplotlib v0.10.1.
\begin{tikzpicture}

\definecolor{darkgray176}{RGB}{176,176,176}
\definecolor{dodgerblue}{RGB}{30,144,255}
\definecolor{lightgray204}{RGB}{204,204,204}

\begin{axis}[
legend cell align={left},
legend style={
  fill opacity=0.8,
  draw opacity=1,
  text opacity=1,
  at={(0.03,0.97)},
  anchor=north west,
  draw=lightgray204
},
tick align=outside,
tick pos=left,
title={Feature effect of \(\displaystyle X_{\mathtt{month}}\)},
x grid style={darkgray176},
xlabel={\(\displaystyle X_{\mathtt{month}}\)},
xmin=0.45, xmax=12.55,
xtick style={color=black},
y grid style={darkgray176},
ylabel={\(\displaystyle Y_{\mathtt{counts}}\)},
ymin=-29.183798458369, ymax=14.7468616660552,
ytick style={color=black}
]
\addplot [semithick, black]
table {%
1 -27.1869506835938
1.17617619037628 -25.871150970459
1.29729723930359 -24.9201946258545
1.40740740299225 -24.0201549530029
1.51751756668091 -23.0867099761963
1.63863861560822 -22.0221939086914
1.74874877929688 -21.0186061859131
1.85885882377625 -19.9816131591797
1.9689689874649 -18.9112167358398
2.09009003639221 -17.6963996887207
2.32132124900818 -15.3969707489014
2.64064073562622 -12.2988424301147
2.87187194824219 -10.1101322174072
3.1141140460968 -7.87898731231689
3.24624633789062 -6.75733757019043
3.35635638237 -5.88665103912354
3.45545554161072 -5.15343141555786
3.56556558609009 -4.3964409828186
3.67567563056946 -3.69928789138794
3.77477478981018 -3.12166857719421
3.88488483428955 -2.53821158409119
3.99499487876892 -2.01459169387817
4.13813829421997 -1.40747404098511
4.23723745346069 -1.0118283033371
4.34734725952148 -0.59662914276123
4.45745754241943 -0.207057595252991
4.56756734848022 0.156886339187622
4.66666650772095 0.462822675704956
4.7767767906189 0.778071641921997
4.88688707351685 1.06769299507141
4.98598575592041 1.30698728561401
5.21721744537354 1.81831657886505
5.43743753433228 2.28735709190369
5.64664649963379 2.71674489974976
5.86686706542969 3.15148568153381
6.12012004852295 3.64487433433533
6.19719696044922 3.81250214576721
6.30730724334717 4.07315158843994
6.42842864990234 4.38769626617432
6.53853845596313 4.69808197021484
6.64864873886108 5.03215312957764
6.75875854492188 5.3899097442627
6.87987995147705 5.81102657318115
6.98998975753784 6.21852016448975
7.23223209381104 7.15501070022583
7.463463306427 8.071533203125
7.6836838722229 8.96508026123047
7.90390396118164 9.87869548797607
8.09109115600586 10.6491012573242
8.17917919158936 10.9760904312134
8.27827835083008 11.3111982345581
8.38838863372803 11.6416292190552
8.49849891662598 11.9293622970581
8.60860824584961 12.1743965148926
8.70770740509033 12.3600397109985
8.81781768798828 12.5239429473877
8.92792797088623 12.6451482772827
9.03803825378418 12.7230014801025
9.1371374130249 12.7500133514404
9.17016983032227 12.7448768615723
9.24724769592285 12.7297611236572
9.28028011322021 12.7090911865234
9.35735702514648 12.6578493118286
9.40140151977539 12.6091642379761
9.47847843170166 12.5178241729736
9.58858871459961 12.3373823165894
9.69869899749756 12.1052808761597
9.80880928039551 11.8215208053589
9.92992973327637 11.4478168487549
10.1061058044434 10.8241405487061
10.2162160873413 10.4159374237061
10.3263263702393 9.99349880218506
10.4364366531372 9.55682373046875
10.5575571060181 9.06051921844482
10.667667388916 8.59395122528076
10.777777671814 8.11314868927002
10.8988990783691 7.56844854354858
11.7247247695923 3.76278614997864
12 2.50704598426819
};
\addlegendentry{$f_{\mathtt{DALE}}$}
\addplot [semithick, dodgerblue, dashed]
table {%
1 -27.0010604858398
1.18718719482422 -25.586950302124
1.29729723930359 -24.7142314910889
1.40740740299225 -23.809778213501
1.52852857112885 -22.77903175354
1.63863861560822 -21.8079357147217
1.74874877929688 -20.8051052093506
1.86986982822418 -19.666467666626
1.99099099636078 -18.4894695281982
2.2992992401123 -15.4285402297974
2.51951956748962 -13.2975368499756
2.72872877120972 -11.3166017532349
2.94894886016846 -9.27656745910645
3.1141140460968 -7.78738117218018
3.24624633789062 -6.68413352966309
3.35635638237 -5.82812023162842
3.45545554161072 -5.10758543014526
3.56556558609009 -4.36409330368042
3.67567563056946 -3.67981958389282
3.77477478981018 -3.113276720047
3.88488483428955 -2.54152417182922
3.99499487876892 -2.02899026870728
4.13813829421997 -1.43530428409576
4.23723745346069 -1.04830312728882
4.34734725952148 -0.642061471939087
4.45745754241943 -0.260767459869385
4.56756734848022 0.0955789089202881
4.66666650772095 0.395250797271729
4.7767767906189 0.704194068908691
4.88688707351685 0.988189816474915
4.98598575592041 1.22298789024353
5.23923921585083 1.77311503887177
5.4484486579895 2.21300554275513
5.66866874694824 2.66170525550842
5.88888883590698 3.09569883346558
6.10910892486572 3.52712678909302
6.19719696044922 3.71933627128601
6.30730724334717 3.9817156791687
6.42842864990234 4.29802989959717
6.53853845596313 4.60990715026855
6.64864873886108 4.9453558921814
6.75875854492188 5.30437612533569
6.87987995147705 5.72675085067749
6.98998975753784 6.13526916503906
7.24324321746826 7.11681127548218
7.463463306427 7.99168014526367
7.6836838722229 8.8862886428833
7.90390396118164 9.80063629150391
8.09109115600586 10.5716905593872
8.17917919158936 10.899432182312
8.2892894744873 11.2697401046753
8.38838863372803 11.5681610107422
8.49849891662598 11.8583631515503
8.60860824584961 12.106406211853
8.70770740509033 12.2951984405518
8.81781768798828 12.463134765625
8.92792797088623 12.5889139175415
9.03803825378418 12.6718053817749
9.1371374130249 12.702956199646
9.17016983032227 12.6990699768066
9.24724769592285 12.6868410110474
9.28028011322021 12.6672773361206
9.35735702514648 12.6185913085938
9.40140151977539 12.5711889266968
9.47847843170166 12.4820346832275
9.58858871459961 12.3042573928833
9.69869899749756 12.0743465423584
9.80880928039551 11.7923011779785
9.92992973327637 11.4199190139771
10.1061058044434 10.7974834442139
10.2162160873413 10.3899793624878
10.3263263702393 9.96817970275879
10.4364366531372 9.53208541870117
10.5575571060181 9.03635120391846
10.667667388916 8.57023906707764
10.777777671814 8.08983135223389
10.8988990783691 7.54549980163574
11.7027025222778 3.84338307380676
12 2.48848509788513
};
\addlegendentry{$\hat{f}_{\mathtt{ALE}}$}
\end{axis}

\end{tikzpicture}}
  \resizebox{.3\columnwidth}{!}{% This file was created with tikzplotlib v0.10.1.
\begin{tikzpicture}

\definecolor{darkgray176}{RGB}{176,176,176}
\definecolor{dodgerblue}{RGB}{30,144,255}
\definecolor{lightgray204}{RGB}{204,204,204}

\begin{axis}[
legend cell align={left},
legend style={
  fill opacity=0.8,
  draw opacity=1,
  text opacity=1,
  at={(0.03,0.97)},
  anchor=north west,
  draw=lightgray204
},
tick align=outside,
tick pos=left,
title={Feature effect of \(\displaystyle X_{\mathtt{atemp}}\)},
x grid style={darkgray176},
xlabel={\(\displaystyle X_{\mathtt{atemp}}\)},
xmin=-19.3, xmax=53.3,
xtick style={color=black},
y grid style={darkgray176},
ylabel={\(\displaystyle Y_{\mathtt{counts}}\)},
ymin=-31.6301535780892, ymax=23.4795271752803,
ytick style={color=black}
]
\addplot [semithick, black]
table {%
-16 -29.1201610565186
-14.5465469360352 -29.1236610412598
-13.3573570251465 -29.1187324523926
-12.6306304931641 -28.9887657165527
-11.9699697494507 -28.7167987823486
-11.3753757476807 -28.351110458374
-10.5825824737549 -28.10817527771
-9.9219217300415 -27.8089466094971
-9.26126098632812 -27.4157428741455
-8.13813781738281 -26.5979690551758
-6.94894886016846 -25.7284774780273
-6.55255270004272 -25.4859237670898
-5.89189195632935 -25.1317119598389
-4.70270252227783 -24.557861328125
-3.90990996360779 -24.0795192718506
-3.11711716651917 -23.5240688323975
-2.52252244949341 -23.0663394927979
-1.86186182498932 -22.5092353820801
-0.540540456771851 -21.3122959136963
0.120120167732239 -20.6509094238281
0.714714765548706 -20.0558643341064
1.30930936336517 -19.4900760650635
2.03603601455688 -18.7049884796143
2.63063073158264 -17.9895362854004
3.8858859539032 -16.3134365081787
5.27327346801758 -14.3682708740234
6 -13.2194204330444
6.85885906219482 -11.8082704544067
7.25525522232056 -11.203857421875
7.91591596603394 -10.3020009994507
9.30330371856689 -8.58228397369385
9.83183193206787 -7.77181529998779
10.9549551010132 -5.72904586791992
12.342342376709 -3.19349336624146
12.5405406951904 -2.89349699020386
13.7957954406738 -1.05008816719055
14.4564561843872 -0.0256915092468262
15.6456460952759 2.00035262107849
16.5705699920654 3.49347949028015
17.6936931610107 5.24578619003296
19.9399394989014 8.44653511047363
20.4024028778076 9.01027202606201
21.0630626678467 9.66129207611084
21.7237243652344 10.1941547393799
24.8948955535889 13.3221054077148
25.687686920166 14.134651184082
26.4804801940918 14.7308025360107
27.1411418914795 15.1683645248413
29.8498497009277 16.8362903594971
30.7087078094482 17.4392356872559
31.3693695068359 17.8697376251221
32.3603591918945 18.484224319458
33.087085723877 18.8724536895752
33.5495491027832 19.0880432128906
34.4084091186523 19.3274822235107
35.0690689086914 19.4596900939941
35.5315322875977 19.543493270874
36.9849853515625 20.1296157836914
37.4474487304688 20.3274917602539
37.6456451416016 20.3095550537109
38.1741752624512 20.2521419525146
38.8348350524902 20.287296295166
39.8258247375488 20.4710922241211
40.6186180114746 20.6114158630371
41.345344543457 20.7675323486328
41.4114112854004 20.7821464538574
41.8078079223633 20.7240734100342
42.1381378173828 20.6876392364502
42.7987976074219 20.7667903900146
43.3933944702148 20.9745426177979
43.657657623291 20.8512020111084
44.714714050293 20.3551597595215
45.4414405822754 20.057201385498
46.1021003723145 19.9324207305908
46.7627639770508 19.9523639678955
47.4894905090332 20.1014881134033
48.1501502990723 20.1500225067139
48.8108100891113 20.1130447387695
49.4054069519043 20.0048065185547
50 19.7889881134033
};
\addlegendentry{$f_{\mathtt{DALE}}$}
\addplot [semithick, dodgerblue, dashed]
table {%
-16 -28.9542388916016
-14.5465469360352 -28.9578762054443
-13.3573570251465 -28.9526977539062
-12.6306304931641 -28.8202018737793
-11.9699697494507 -28.5466842651367
-11.3753757476807 -28.1803741455078
-10.6486482620239 -27.9906787872314
-9.98798751831055 -27.7104911804199
-9.26126098632812 -27.2779521942139
-8.00600624084473 -26.3645648956299
-7.2132134437561 -25.7814712524414
-6.55255270004272 -25.3670864105225
-5.89189195632935 -25.0190086364746
-4.76876878738403 -24.4896812438965
-3.90990996360779 -23.9704494476318
-3.05105113983154 -23.364631652832
-2.52252244949341 -22.9567375183105
-1.86186182498932 -22.4021339416504
-0.606606602668762 -21.2720623016357
0.120120167732239 -20.5483741760254
0.648648619651794 -20.0188083648682
1.30930936336517 -19.4072608947754
1.96996998786926 -18.7101631164551
2.69669675827026 -17.8467922210693
3.55555558204651 -16.7303333282471
4.21621608734131 -15.8174858093262
5.53753757476807 -13.9267387390137
6 -13.1932601928711
6.66066074371338 -12.1299390792847
7.32132148742676 -11.1632461547852
7.91591596603394 -10.3738870620728
8.97297286987305 -9.1130838394165
9.10510540008545 -8.9506664276123
9.83183193206787 -7.83912706375122
11.2852849960327 -5.17018270492554
12.408408164978 -3.13905572891235
13.5975971221924 -1.41476833820343
14.3243246078491 -0.306623458862305
14.6546545028687 0.264089107513428
15.6456460952759 1.96240651607513
16.5705699920654 3.46715188026428
17.6936931610107 5.24065256118774
19.8738746643066 8.35614967346191
20.4024028778076 9.0117301940918
21.0630626678467 9.66457653045654
21.7237243652344 10.1938533782959
25.4894886016846 13.9263477325439
25.6216220855713 14.053521156311
26.4804801940918 14.7064094543457
27.1411418914795 15.1473188400269
29.057056427002 16.3006134033203
29.6516513824463 16.6701507568359
30.6426429748535 17.3715686798096
31.2372379302979 17.7557601928711
32.4924926757812 18.524055480957
33.087085723877 18.8392581939697
33.5495491027832 19.0558605194092
34.4084091186523 19.2951412200928
35.0690689086914 19.4283008575439
35.5315322875977 19.5130195617676
36.9849853515625 20.0959377288818
37.4474487304688 20.291877746582
37.6456451416016 20.2738914489746
38.1741752624512 20.216480255127
38.8348350524902 20.2516288757324
39.8258247375488 20.4354114532471
40.6186180114746 20.5759296417236
41.2792778015137 20.7178859710693
41.4114112854004 20.7472190856934
41.8078079223633 20.6907978057861
42.1381378173828 20.6556282043457
42.7987976074219 20.7358493804932
43.3933944702148 20.9432563781738
43.657657623291 20.8323345184326
45.375373840332 20.0885391235352
46.0360374450684 19.9492740631104
46.6966972351074 19.9548187255859
47.5555572509766 20.1251811981201
48.1501502990723 20.1673374176025
48.8108100891113 20.1304454803467
49.4054069519043 20.0222206115723
50 19.8063983917236
};
\addlegendentry{$\hat{f}_{\mathtt{ALE}}$}
\end{axis}

\end{tikzpicture}}
  \resizebox{.3\columnwidth}{!}{% This file was created with tikzplotlib v0.10.1.
\begin{tikzpicture}

\definecolor{darkgray176}{RGB}{176,176,176}
\definecolor{dodgerblue}{RGB}{30,144,255}
\definecolor{lightgray204}{RGB}{204,204,204}

\begin{axis}[
legend cell align={left},
legend style={fill opacity=0.8, draw opacity=1, text opacity=1, draw=lightgray204},
tick align=outside,
tick pos=left,
title={Feature effect of \(\displaystyle X_{\mathtt{hum}}\)},
x grid style={darkgray176},
xlabel={\(\displaystyle X_{\mathtt{hum}}\)},
xmin=-5, xmax=105,
xtick style={color=black},
y grid style={darkgray176},
ylabel={\(\displaystyle Y_{\mathtt{counts}}\)},
ymin=-12.4017583163331, ymax=20.9680093156312,
ytick style={color=black}
]
\addplot [semithick, black]
table {%
0 19.445671081543
1.80180180072784 19.2508811950684
2.8028028011322 19.1621017456055
3.80380392074585 19.0876598358154
4.80480480194092 19.0275573730469
5.80580568313599 18.9817924499512
6.80680704116821 18.9503650665283
7.80780792236328 18.933277130127
8.80880928039551 18.9305267333984
9.00900936126709 18.9289054870605
10.010009765625 18.7470264434814
11.0110111236572 18.3885173797607
12.0120124816895 18.1801509857178
13.0130128860474 18.1236801147461
14.0140142440796 18.3442001342773
15.0150146484375 18.2827415466309
16.0160160064697 18.0843563079834
17.1171169281006 17.4235782623291
18.0180187225342 16.9714221954346
19.0190181732178 16.268253326416
20.02001953125 15.9827404022217
21.2212219238281 15.4168157577515
23.2232227325439 14.5354175567627
24.4244251251221 13.9460258483887
25.325325012207 13.5335330963135
26.2262268066406 13.1304426193237
27.027027130127 12.738260269165
28.0280284881592 12.1465797424316
29.0290298461914 11.7472257614136
30.030029296875 11.1805286407471
31.9319324493408 10.4186792373657
32.1321334838867 10.3230867385864
33.033031463623 9.86366176605225
34.1341323852539 9.40140438079834
36.9369354248047 8.0537805557251
37.4374389648438 7.83730363845825
39.2392387390137 7.05814123153687
40.1401405334473 6.63687181472778
41.5415420532227 6.06709289550781
42.5425415039062 5.68619441986084
43.3433418273926 5.3793511390686
44.1441459655762 5.07060098648071
45.1451454162598 4.7441782951355
46.2462463378906 4.34106254577637
50.1501502990723 3.03945064544678
51.1511497497559 2.67909049987793
52.352352142334 2.33571791648865
53.1531524658203 2.11605978012085
54.5545539855957 1.6564826965332
55.1551551818848 1.46061563491821
56.6566581726074 1.04154002666473
58.0580596923828 0.674215078353882
59.4594612121582 0.312742352485657
60.4604606628418 0.0281145572662354
61.1611595153809 -0.170746326446533
62.0620613098145 -0.390427708625793
63.1631622314453 -0.749048829078674
64.1641616821289 -1.0204029083252
65.0650634765625 -1.29337751865387
66.1661682128906 -1.49768507480621
67.0670700073242 -1.71112847328186
68.0680694580078 -2.03258061408997
69.0690689086914 -2.23863196372986
70.070068359375 -2.38389420509338
71.0710678100586 -2.61734461784363
73.3733749389648 -3.33057689666748
74.1741714477539 -3.53978538513184
75.175178527832 -3.74607062339783
75.9759750366211 -3.88061666488647
76.3763732910156 -4.00896739959717
77.1771774291992 -4.26187658309937
78.2782745361328 -4.54429483413696
79.0790786743164 -4.72437953948975
80.1801834106445 -4.89549112319946
80.9809799194336 -4.98908805847168
81.3813781738281 -5.09695863723755
82.8828811645508 -5.50094985961914
83.5835800170898 -5.67414093017578
84.584587097168 -5.90682792663574
85.185188293457 -6.05287551879883
85.9859848022461 -6.28149127960205
86.4864883422852 -6.34115266799927
87.0870895385742 -6.42070388793945
88.6886901855469 -6.81395483016968
90.9909896850586 -7.4021258354187
91.2912902832031 -7.73944664001465
91.9919891357422 -8.54510974884033
92.1921920776367 -8.59602737426758
94.0940933227539 -9.01366710662842
94.9949951171875 -9.32887077331543
95.9959945678711 -9.81900787353516
96.9969940185547 -10.4498987197876
97.7977981567383 -10.5560169219971
98.7987976074219 -10.6977682113647
99.7997970581055 -10.85329246521
100 -10.8849506378174
};
\addlegendentry{$f_{\mathtt{DALE}}$}
\addplot [semithick, dodgerblue, dashed]
table {%
0 19.4512023925781
1.80180180072784 19.2567539215088
2.8028028011322 19.1681327819824
3.80380392074585 19.093822479248
4.80480480194092 19.0338249206543
5.80580568313599 18.9881401062012
6.80680704116821 18.9567699432373
7.80780792236328 18.9397106170654
8.80880928039551 18.9369659423828
9.00900936126709 18.9353446960449
10.010009765625 18.753490447998
11.0110111236572 18.3950309753418
12.0120124816895 18.1866893768311
13.0130128860474 18.1302185058594
14.0140142440796 18.3507404327393
15.0150146484375 18.289701461792
16.0160160064697 18.0913181304932
17.1171169281006 17.4305477142334
18.0180187225342 16.9784984588623
19.0190181732178 16.2804527282715
20.02001953125 15.9955158233643
21.2212219238281 15.4311323165894
23.1231231689453 14.5908050537109
24.3243236541748 13.9988222122192
25.325325012207 13.5421257019043
26.1261253356934 13.187403678894
27.027027130127 12.7418575286865
28.0280284881592 12.1519727706909
29.0290298461914 11.7499494552612
30.030029296875 11.182975769043
32.0320320129395 10.3707160949707
33.033031463623 9.85312747955322
34.1341323852539 9.38883113861084
36.6366348266602 8.18614196777344
37.1371383666992 7.95634317398071
39.0390396118164 7.15220594406128
40.1401405334473 6.63696813583374
41.4414405822754 6.10159015655518
42.4424438476562 5.71701908111572
43.3433418273926 5.37234306335449
44.1441459655762 5.06235647201538
45.1451454162598 4.73639726638794
46.2462463378906 4.33528661727905
50.050048828125 3.06395077705383
51.1511497497559 2.66571760177612
52.2522506713867 2.34696483612061
53.1531524658203 2.10619187355042
54.5545539855957 1.64742040634155
55.1551551818848 1.4514753818512
56.5565567016602 1.06200659275055
58.1581573486328 0.648456811904907
59.4594612121582 0.313627243041992
60.4604606628418 0.0284990072250366
61.1611595153809 -0.170641422271729
62.0620613098145 -0.391844630241394
63.0630645751953 -0.725825071334839
64.2642669677734 -1.05191004276276
65.0650634765625 -1.28894066810608
66.1661682128906 -1.49367189407349
67.0670700073242 -1.70696103572845
68.0680694580078 -2.03458189964294
69.0690689086914 -2.24001359939575
70.070068359375 -2.38448262214661
71.0710678100586 -2.62038135528564
73.2732696533203 -3.31713366508484
74.1741714477539 -3.55359435081482
75.175178527832 -3.7575466632843
75.9759750366211 -3.88404536247253
76.2762756347656 -3.97914242744446
77.1771774291992 -4.2657265663147
78.2782745361328 -4.54978799819946
79.1791763305664 -4.74476099014282
80.1801834106445 -4.89946508407593
80.9809799194336 -4.99041557312012
81.3813781738281 -5.09758758544922
82.9829864501953 -5.52745628356934
84.2842864990234 -5.83696460723877
85.185188293457 -6.04725694656372
85.9859848022461 -6.27391958236694
86.4864883422852 -6.33382940292358
87.0870895385742 -6.41337013244629
88.6886901855469 -6.80084180831909
90.9909896850586 -7.36907291412354
91.2912902832031 -7.70228242874146
91.9919891357422 -8.49856758117676
92.1921920776367 -8.5493803024292
94.0940933227539 -8.96699905395508
94.9949951171875 -9.28218460083008
95.9959945678711 -9.77230548858643
96.9969940185547 -10.4031848907471
97.7977981567383 -10.509295463562
98.5986022949219 -10.6244306564331
99.5996017456055 -10.7833385467529
100 -10.8497257232666
};
\addlegendentry{$\hat{f}_{\mathtt{ALE}}$}
\end{axis}

\end{tikzpicture}}
  \caption{Bike-Sharing Dataset. DALE and ALE feature effect plots with \(K=200\) for: \(X_{\texttt{month}}\), \(X_{\mathtt{atemp}}\),\(X_{\mathtt{hum}}\).}
  \label{fig:bike-sharing-comparison}
\end{figure}

In the case of the Bike-Sharing, it is infeasible compare to compute the ground-truth ALE. We have lack of knowledge about the data-generating distribution and the dimensionality of the problem \(D=11\) is prohibitive for applying numerical integration on Eq.~\eqref{eq:ALE}. However, given the fact that the dataset has a large number of instances, DALE and ALE provide almost identical approximations for all features for large \(K\) as we confirm in Figure~\ref{fig:bike-sharing-comparison}. We also notice the for all feature features, except \(X_{\mathtt{hour}}\), lowering the number of bins \(K\) does not significantly impacts the approximation, since these features change slowly wrt the feature value.

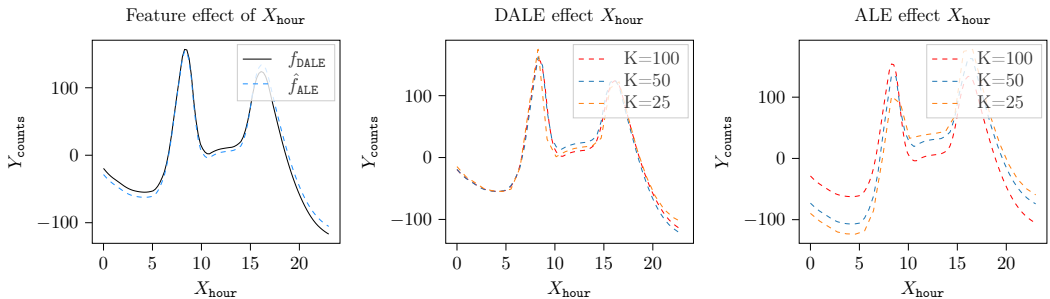
\begin{figure}[h]
  \centering
  \resizebox{.3\columnwidth}{!}{% This file was created with tikzplotlib v0.10.1.
\begin{tikzpicture}

\definecolor{darkgray176}{RGB}{176,176,176}
\definecolor{dodgerblue}{RGB}{30,144,255}
\definecolor{lightgray204}{RGB}{204,204,204}

\begin{axis}[
legend cell align={left},
legend style={fill opacity=0.8, draw opacity=1, text opacity=1, draw=lightgray204},
tick align=outside,
tick pos=left,
title={Feature effect of \(\displaystyle X_{\mathtt{hour}}\)},
x grid style={darkgray176},
xlabel={\(\displaystyle X_{\mathtt{hour}}\)},
xmin=-1.15, xmax=24.15,
xtick style={color=black},
y grid style={darkgray176},
ylabel={\(\displaystyle Y_{\mathtt{counts}}\)},
ymin=-130.55351264475, ymax=170.472616656964,
ytick style={color=black}
]
\addplot [semithick, black]
table {%
0 -19.5050849914551
0.276276350021362 -24.3441772460938
0.506506443023682 -27.9236030578613
0.73673677444458 -31.0645523071289
0.943943977355957 -33.5316467285156
1.54254257678986 -39.7169952392578
2.00300312042236 -44.5988159179688
2.11811804771423 -45.7644081115723
2.32532525062561 -47.663459777832
2.55555558204651 -49.4630393981934
2.78578567504883 -50.9483070373535
2.99299287796021 -52.0300178527832
3.24624633789062 -52.9924545288086
3.47647643089294 -53.6967239379883
3.68368363380432 -54.1934471130371
3.91391396522522 -54.5767364501953
4.14414405822754 -54.7889671325684
4.37437438964844 -54.7139015197754
4.60460472106934 -54.3515357971191
4.83483505249023 -53.7018775939941
5.04204225540161 -52.8642959594727
5.06506490707397 -52.7370185852051
5.29529523849487 -50.2154121398926
5.52552556991577 -46.1370544433594
5.75575590133667 -40.5019493103027
6.00900888442993 -32.4394378662109
6.21621608734131 -24.4946517944336
6.44644641876221 -11.6443128585815
6.67667675018311 5.24092245101929
6.92992973327637 28.6440601348877
7.43643665313721 82.6207962036133
7.89689683914185 130.067947387695
8.05805778503418 145.936157226562
8.26526546478271 156.092514038086
8.28828811645508 156.789611816406
8.49549579620361 156.161087036133
8.51851844787598 155.647872924805
8.72572612762451 144.234481811523
8.74874877929688 142.510955810547
8.97897911071777 117.378860473633
9.23223209381104 78.1012954711914
9.43943977355957 52.0760345458984
9.66967010498047 30.5269412994385
9.89989948272705 16.3793964385986
10.1301298141479 9.40791988372803
10.3603601455688 4.70836496353149
10.5675678253174 2.42953133583069
10.5905904769897 2.28073143959045
10.7977981567383 2.04458355903625
10.8208208084106 2.12501859664917
11.0740737915039 4.29217100143433
11.3043041229248 5.98237943649292
11.5345344543457 7.39637041091919
11.7417421340942 8.44379425048828
11.9719715118408 9.33294486999512
12.2482481002808 10.0710592269897
12.4784784317017 10.624490737915
12.7087087631226 11.1187515258789
12.9159154891968 11.5147294998169
13.1001005172729 11.8276166915894
13.1231231689453 11.9058475494385
13.3303298950195 12.875524520874
13.3763761520386 13.199444770813
13.5835838317871 14.8272914886475
13.813814163208 17.3185062408447
14.0210208892822 20.1412181854248
14.0670671463013 21.3190975189209
14.2512512207031 26.5641231536865
14.2972974777222 28.4294090270996
14.5045042037964 37.6234588623047
14.7347345352173 51.2384872436523
14.9649648666382 68.2768707275391
15.1951951980591 88.2222595214844
15.42542552948 103.759735107422
15.6556558609009 114.889282226562
15.8628625869751 121.212203979492
15.8858861923218 121.610908508301
16.0930938720703 123.970664978027
16.1161155700684 123.936637878418
16.3233242034912 122.487854003906
16.3693695068359 121.442276000977
16.5535526275635 116.769096374512
16.6226234436035 113.865631103516
16.7837829589844 106.814399719238
16.8758754730225 101.206695556641
17.3363361358643 72.9654541015625
17.7967967987061 45.5679664611816
18.2112121582031 21.7020950317383
18.5105113983154 5.64360857009888
18.7407398223877 -5.97945785522461
18.9709701538086 -16.9456024169922
19.1781787872314 -26.2628364562988
19.4084091186523 -35.9988479614258
19.6386394500732 -45.0910110473633
19.8688697814941 -53.5393295288086
20.0990982055664 -61.3896484375
20.3293285369873 -68.7146987915039
20.5365371704102 -74.8665542602539
20.7667675018311 -81.1935348510742
20.996997833252 -86.995246887207
21.2042045593262 -91.7762222290039
21.4344348907471 -96.5489959716797
21.664665222168 -100.770584106445
21.8948955535889 -104.44100189209
22.1481475830078 -107.900184631348
22.3553562164307 -110.46851348877
22.5855846405029 -113.02222442627
22.8158149719238 -115.269233703613
23 -116.870506286621
};
\addlegendentry{$f_{\mathtt{DALE}}$}
\addplot [semithick, dodgerblue, dashed]
table {%
0 -28.6461296081543
0.299299240112305 -33.4844398498535
0.506506443023682 -36.5122833251953
0.73673677444458 -39.5242080688477
0.966966986656189 -42.1757049560547
1.74974977970123 -50.2825317382812
2.11811804771423 -54.0799827575684
2.32532525062561 -55.9247512817383
2.55555558204651 -57.6483955383301
2.78578567504883 -59.0419044494629
2.99299287796021 -60.0281982421875
3.24624633789062 -60.8593711853027
3.47647643089294 -61.46630859375
3.68368363380432 -61.8931121826172
3.91391396522522 -62.2204475402832
4.14414405822754 -62.3986968994141
4.37437438964844 -62.3219032287598
4.60460472106934 -61.990062713623
4.83483505249023 -61.4031791687012
5.04204225540161 -60.6504898071289
5.06506490707397 -60.5277214050293
5.29529523849487 -57.8714256286621
5.52552556991577 -53.4342956542969
5.75575590133667 -47.2163352966309
6.00900888442993 -38.2443809509277
6.21621608734131 -29.381742477417
6.44644641876221 -15.6827802658081
6.69969987869263 4.01029825210571
6.92992973327637 25.8215217590332
7.39039039611816 75.5200271606445
7.85085105895996 123.192329406738
8.05805778503418 143.596130371094
8.26526546478271 153.462203979492
8.28828811645508 154.119873046875
8.49549579620361 153.021850585938
8.51851844787598 152.449096679688
8.72572612762451 140.386978149414
8.74874877929688 138.58381652832
8.97897911071777 112.524017333984
9.20920944213867 75.0672760009766
9.43943977355957 45.3751182556152
9.66967010498047 23.4475517272949
9.89989948272705 9.28457546234131
10.1301298141479 2.64411807060242
10.3603601455688 -1.73881900310516
10.5675678253174 -3.74506211280823
10.5905904769897 -3.86423587799072
10.7977981567383 -3.84074091911316
10.8208208084106 -3.73213291168213
11.0740737915039 -1.27214574813843
11.3043041229248 0.624772906303406
11.5115118026733 2.059002161026
11.7417421340942 3.31951761245728
11.9719715118408 4.24510145187378
12.2482481002808 4.96000051498413
12.4784784317017 5.49322175979614
12.7087087631226 5.96644258499146
12.9159154891968 6.34280109405518
13.1001005172729 6.63770151138306
13.1231231689453 6.70726346969604
13.3303298950195 7.55532884597778
13.3763761520386 7.83442735671997
13.5835838317871 9.23264122009277
13.813814163208 11.3566951751709
14.0210208892822 13.7536668777466
14.0440444946289 14.2715625762939
14.2512512207031 20.38014793396
14.2972974777222 22.4057006835938
14.5045042037964 32.4857139587402
14.7347345352173 47.7849960327148
14.9649648666382 67.2124176025391
15.1951951980591 90.178352355957
15.42542552948 108.329772949219
15.6556558609009 121.666687011719
15.8628625869751 129.635528564453
15.9089088439941 130.593536376953
16.0930938720703 133.829193115234
16.1161155700684 133.91423034668
16.3233242034912 133.447540283203
16.3463459014893 133.071701049805
16.5535526275635 128.498184204102
16.5995998382568 126.696998596191
16.7837829589844 118.981163024902
16.8758754730225 113.421195983887
17.3363361358643 85.3754196166992
17.7967967987061 58.1768074035645
18.2112121582031 34.4831466674805
18.5335330963135 17.2674236297607
18.7407398223877 6.80826711654663
18.9709701538086 -4.24147033691406
19.2012004852295 -14.6806631088257
19.4084091186523 -23.5517692565918
19.6386394500732 -32.8170890808105
19.8688697814941 -41.4646186828613
20.0990982055664 -49.5049858093262
20.3063068389893 -56.2461433410645
20.5365371704102 -63.1648941040039
20.7667675018311 -69.4933547973633
20.9739742279053 -74.6995239257812
21.2272281646729 -80.4064331054688
21.4344348907471 -84.651123046875
21.664665222168 -88.8804550170898
21.8948955535889 -92.6121520996094
22.1481475830078 -96.1844940185547
22.3553562164307 -98.8407440185547
22.5855846405029 -101.486793518066
22.8158149719238 -103.820686340332
23 -105.487968444824
};
\addlegendentry{$\hat{f}_{\mathtt{ALE}}$}
\end{axis}

\end{tikzpicture}}
  \resizebox{.3\columnwidth}{!}{% This file was created with tikzplotlib v0.10.1.
\begin{tikzpicture}

\definecolor{darkgray176}{RGB}{176,176,176}
\definecolor{darkorange25512714}{RGB}{255,127,14}
\definecolor{lightgray204}{RGB}{204,204,204}
\definecolor{steelblue31119180}{RGB}{31,119,180}

\begin{axis}[
legend cell align={left},
legend style={fill opacity=0.8, draw opacity=1, text opacity=1, draw=lightgray204},
tick align=outside,
tick pos=left,
title={DALE effect \(\displaystyle X_{\mathtt{hour}}\)},
x grid style={darkgray176},
xlabel={\(\displaystyle X_{\mathtt{hour}}\)},
xmin=-1.15, xmax=24.15,
xtick style={color=black},
y grid style={darkgray176},
ylabel={\(\displaystyle Y_{\mathtt{counts}}\)},
ymin=-138.606064986267, ymax=189.589636529361,
ytick style={color=black}
]
\addplot [semithick, red, dashed]
table {%
0 -19.5050849914551
0.276276350021362 -24.3441772460938
0.506506443023682 -27.9236030578613
0.73673677444458 -31.0645523071289
0.943943977355957 -33.5316467285156
1.54254257678986 -39.7169952392578
2.00300312042236 -44.5988159179688
2.11811804771423 -45.7644081115723
2.32532525062561 -47.663459777832
2.55555558204651 -49.4630393981934
2.78578567504883 -50.9483070373535
2.99299287796021 -52.0300178527832
3.24624633789062 -52.9924545288086
3.47647643089294 -53.6967239379883
3.68368363380432 -54.1934471130371
3.91391396522522 -54.5767364501953
4.14414405822754 -54.7889671325684
4.37437438964844 -54.7139015197754
4.60460472106934 -54.3515357971191
4.83483505249023 -53.7018775939941
5.04204225540161 -52.8642959594727
5.06506490707397 -52.7370185852051
5.29529523849487 -50.2154121398926
5.52552556991577 -46.1370544433594
5.75575590133667 -40.5019493103027
6.00900888442993 -32.4394378662109
6.21621608734131 -24.4946517944336
6.44644641876221 -11.6443128585815
6.67667675018311 5.24092245101929
6.92992973327637 28.6440601348877
7.43643665313721 82.6207962036133
7.89689683914185 130.067947387695
8.05805778503418 145.936157226562
8.26526546478271 156.092514038086
8.28828811645508 156.789611816406
8.49549579620361 156.161087036133
8.51851844787598 155.647872924805
8.72572612762451 144.234481811523
8.74874877929688 142.510955810547
8.97897911071777 117.378860473633
9.23223209381104 78.1012954711914
9.43943977355957 52.0760345458984
9.66967010498047 30.5269412994385
9.89989948272705 16.3793964385986
10.1301298141479 9.40791988372803
10.3603601455688 4.70836496353149
10.5675678253174 2.42953133583069
10.5905904769897 2.28073143959045
10.7977981567383 2.04458355903625
10.8208208084106 2.12501859664917
11.0740737915039 4.29217100143433
11.3043041229248 5.98237943649292
11.5345344543457 7.39637041091919
11.7417421340942 8.44379425048828
11.9719715118408 9.33294486999512
12.2482481002808 10.0710592269897
12.4784784317017 10.624490737915
12.7087087631226 11.1187515258789
12.9159154891968 11.5147294998169
13.1001005172729 11.8276166915894
13.1231231689453 11.9058475494385
13.3303298950195 12.875524520874
13.3763761520386 13.199444770813
13.5835838317871 14.8272914886475
13.813814163208 17.3185062408447
14.0210208892822 20.1412181854248
14.0670671463013 21.3190975189209
14.2512512207031 26.5641231536865
14.2972974777222 28.4294090270996
14.5045042037964 37.6234588623047
14.7347345352173 51.2384872436523
14.9649648666382 68.2768707275391
15.1951951980591 88.2222595214844
15.42542552948 103.759735107422
15.6556558609009 114.889282226562
15.8628625869751 121.212203979492
15.8858861923218 121.610908508301
16.0930938720703 123.970664978027
16.1161155700684 123.936637878418
16.3233242034912 122.487854003906
16.3693695068359 121.442276000977
16.5535526275635 116.769096374512
16.6226234436035 113.865631103516
16.7837829589844 106.814399719238
16.8758754730225 101.206695556641
17.3363361358643 72.9654541015625
17.7967967987061 45.5679664611816
18.2112121582031 21.7020950317383
18.5105113983154 5.64360857009888
18.7407398223877 -5.97945785522461
18.9709701538086 -16.9456024169922
19.1781787872314 -26.2628364562988
19.4084091186523 -35.9988479614258
19.6386394500732 -45.0910110473633
19.8688697814941 -53.5393295288086
20.0990982055664 -61.3896484375
20.3293285369873 -68.7146987915039
20.5365371704102 -74.8665542602539
20.7667675018311 -81.1935348510742
20.996997833252 -86.995246887207
21.2042045593262 -91.7762222290039
21.4344348907471 -96.5489959716797
21.664665222168 -100.770584106445
21.8948955535889 -104.44100189209
22.1481475830078 -107.900184631348
22.3553562164307 -110.46851348877
22.5855846405029 -113.02222442627
22.8158149719238 -115.269233703613
23 -116.870506286621
};
\addlegendentry{K=100}
\addplot [semithick, steelblue31119180, dashed]
table {%
0 -19.0568523406982
0.483483552932739 -27.5899829864502
0.943943977355957 -33.9586143493652
1.65765762329102 -41.3208885192871
2.11811804771423 -46.2027359008789
2.30230236053467 -48.1648330688477
2.76276278495789 -51.5102844238281
3.2232232093811 -53.2904281616211
3.68368363380432 -54.3948593139648
4.12112092971802 -54.8079147338867
4.14414405822754 -54.8193206787109
4.60460472106934 -54.0945930480957
5.04204225540161 -52.3263702392578
5.06506490707397 -52.1762809753418
5.52552556991577 -45.0688095092773
5.98598575592041 -32.7721710205078
6.44644641876221 -15.1332197189331
6.906907081604 18.6453113555908
7.5515513420105 87.5196990966797
8.01201248168945 134.966445922852
8.26526546478271 160.561309814453
8.28828811645508 161.809616088867
8.72572612762451 149.098709106445
8.74874877929688 147.291305541992
9.20920944213867 74.5869293212891
9.66967010498047 31.4887447357178
10.1301298141479 17.5457916259766
10.5675678253174 12.734920501709
10.5905904769897 12.6905250549316
11.0510511398315 16.6829795837402
11.5115118026733 19.7547092437744
11.9719715118408 21.9057159423828
12.4554557800293 23.2053813934326
12.8928928375244 24.1664962768555
13.3303298950195 24.9096031188965
13.3533535003662 25.0283260345459
13.7907905578613 28.3794097900391
13.8368368148804 28.9517726898193
14.2512512207031 34.5971946716309
14.2742738723755 35.4409217834473
14.7117118835449 57.6446723937988
14.757758140564 61.3840827941895
15.1721725463867 97.8089218139648
15.2182178497314 100.394157409668
15.6326322555542 120.966339111328
15.6556558609009 121.510360717773
16.0930938720703 126.492065429688
16.1161155700684 126.161819458008
16.5535526275635 115.063018798828
16.5995998382568 112.439018249512
17.2672672271729 71.3320007324219
17.7277278900146 43.8025321960449
18.1881885528564 16.976188659668
18.4644641876221 1.44218516349792
18.9019012451172 -21.0025215148926
19.362361907959 -42.0182228088379
19.8228225708008 -60.4585456848145
20.2602596282959 -75.6445541381836
20.7207202911377 -89.0095443725586
21.1811809539795 -99.7675628662109
21.6416416168213 -108.320869445801
22.1021022796631 -114.716354370117
22.5625629425049 -119.885055541992
23 -123.688079833984
};
\addlegendentry{K=50}
\addplot [semithick, darkorange25512714, dashed]
table {%
0 -14.6316518783569
0.920920848846436 -31.048433303833
1.97997999191284 -41.9419441223145
2.76276278495789 -50.2953834533691
3.68368363380432 -53.854320526123
4.58158159255981 -54.7021713256836
4.60460472106934 -54.7009506225586
5.50250244140625 -51.0714378356934
5.52552556991577 -50.791748046875
6.42342329025269 -16.8359642028809
6.44644641876221 -15.5138454437256
7.45945930480957 93.0636596679688
8.26526546478271 174.501815795898
8.28828811645508 174.671646118164
9.18618583679199 31.7446269989014
9.20920944213867 29.2628974914551
10.1071071624756 1.68466913700104
10.1301298141479 1.37699365615845
11.0510511398315 9.37294101715088
11.9719715118408 14.6067152023315
12.8928928375244 17.1037063598633
13.7907905578613 18.6290340423584
13.813814163208 18.8328590393066
14.7117118835449 31.0646076202393
14.7347345352173 32.4726219177246
15.6326322555542 111.393112182617
15.6556558609009 112.218955993652
16.5535526275635 122.44457244873
16.5765762329102 121.48802947998
17.5665664672852 60.373119354248
18.4184188842773 11.5534610748291
19.3393402099609 -30.7352237701416
20.2372379302979 -62.1434631347656
20.3753757476807 -65.432746887207
21.1581573486328 -83.8936080932617
21.2962970733643 -85.8475189208984
22.0790786743164 -96.8198623657227
22.3783779144287 -99.4268341064453
23 -104.831130981445
};
\addlegendentry{K=25}
\end{axis}

\end{tikzpicture}}
  \resizebox{.3\columnwidth}{!}{% This file was created with tikzplotlib v0.10.1.
\begin{tikzpicture}

\definecolor{darkgray176}{RGB}{176,176,176}
\definecolor{darkorange25512714}{RGB}{255,127,14}
\definecolor{lightgray204}{RGB}{204,204,204}
\definecolor{steelblue31119180}{RGB}{31,119,180}

\begin{axis}[
legend cell align={left},
legend style={fill opacity=0.8, draw opacity=1, text opacity=1, draw=lightgray204},
tick align=outside,
tick pos=left,
title={ALE effect \(\displaystyle X_{\mathtt{hour}}\)},
x grid style={darkgray176},
xlabel={\(\displaystyle X_{\mathtt{hour}}\)},
xmin=-1.15, xmax=24.15,
xtick style={color=black},
y grid style={darkgray176},
ylabel={\(\displaystyle Y_{\mathtt{counts}}\)},
ymin=-138.606757724745, ymax=192.972018488682,
ytick style={color=black}
]
\addplot [semithick, red, dashed]
table {%
0 -28.6461296081543
0.299299240112305 -33.4844398498535
0.506506443023682 -36.5122833251953
0.73673677444458 -39.5242080688477
0.966966986656189 -42.1757049560547
1.74974977970123 -50.2825317382812
2.11811804771423 -54.0799827575684
2.32532525062561 -55.9247512817383
2.55555558204651 -57.6483955383301
2.78578567504883 -59.0419044494629
2.99299287796021 -60.0281982421875
3.24624633789062 -60.8593711853027
3.47647643089294 -61.46630859375
3.68368363380432 -61.8931121826172
3.91391396522522 -62.2204475402832
4.14414405822754 -62.3986968994141
4.37437438964844 -62.3219032287598
4.60460472106934 -61.990062713623
4.83483505249023 -61.4031791687012
5.04204225540161 -60.6504898071289
5.06506490707397 -60.5277214050293
5.29529523849487 -57.8714256286621
5.52552556991577 -53.4342956542969
5.75575590133667 -47.2163352966309
6.00900888442993 -38.2443809509277
6.21621608734131 -29.381742477417
6.44644641876221 -15.6827802658081
6.69969987869263 4.01029825210571
6.92992973327637 25.8215217590332
7.39039039611816 75.5200271606445
7.85085105895996 123.192329406738
8.05805778503418 143.596130371094
8.26526546478271 153.462203979492
8.28828811645508 154.119873046875
8.49549579620361 153.021850585938
8.51851844787598 152.449096679688
8.72572612762451 140.386978149414
8.74874877929688 138.58381652832
8.97897911071777 112.524017333984
9.20920944213867 75.0672760009766
9.43943977355957 45.3751182556152
9.66967010498047 23.4475517272949
9.89989948272705 9.28457546234131
10.1301298141479 2.64411807060242
10.3603601455688 -1.73881900310516
10.5675678253174 -3.74506211280823
10.5905904769897 -3.86423587799072
10.7977981567383 -3.84074091911316
10.8208208084106 -3.73213291168213
11.0740737915039 -1.27214574813843
11.3043041229248 0.624772906303406
11.5115118026733 2.059002161026
11.7417421340942 3.31951761245728
11.9719715118408 4.24510145187378
12.2482481002808 4.96000051498413
12.4784784317017 5.49322175979614
12.7087087631226 5.96644258499146
12.9159154891968 6.34280109405518
13.1001005172729 6.63770151138306
13.1231231689453 6.70726346969604
13.3303298950195 7.55532884597778
13.3763761520386 7.83442735671997
13.5835838317871 9.23264122009277
13.813814163208 11.3566951751709
14.0210208892822 13.7536668777466
14.0440444946289 14.2715625762939
14.2512512207031 20.38014793396
14.2972974777222 22.4057006835938
14.5045042037964 32.4857139587402
14.7347345352173 47.7849960327148
14.9649648666382 67.2124176025391
15.1951951980591 90.178352355957
15.42542552948 108.329772949219
15.6556558609009 121.666687011719
15.8628625869751 129.635528564453
15.9089088439941 130.593536376953
16.0930938720703 133.829193115234
16.1161155700684 133.91423034668
16.3233242034912 133.447540283203
16.3463459014893 133.071701049805
16.5535526275635 128.498184204102
16.5995998382568 126.696998596191
16.7837829589844 118.981163024902
16.8758754730225 113.421195983887
17.3363361358643 85.3754196166992
17.7967967987061 58.1768074035645
18.2112121582031 34.4831466674805
18.5335330963135 17.2674236297607
18.7407398223877 6.80826711654663
18.9709701538086 -4.24147033691406
19.2012004852295 -14.6806631088257
19.4084091186523 -23.5517692565918
19.6386394500732 -32.8170890808105
19.8688697814941 -41.4646186828613
20.0990982055664 -49.5049858093262
20.3063068389893 -56.2461433410645
20.5365371704102 -63.1648941040039
20.7667675018311 -69.4933547973633
20.9739742279053 -74.6995239257812
21.2272281646729 -80.4064331054688
21.4344348907471 -84.651123046875
21.664665222168 -88.8804550170898
21.8948955535889 -92.6121520996094
22.1481475830078 -96.1844940185547
22.3553562164307 -98.8407440185547
22.5855846405029 -101.486793518066
22.8158149719238 -103.820686340332
23 -105.487968444824
};
\addlegendentry{K=100}
\addplot [semithick, steelblue31119180, dashed]
table {%
0 -73.0295944213867
0.483483552932739 -80.3262100219727
0.943943977355957 -86.0722503662109
1.63463461399078 -93.052131652832
2.09509515762329 -97.846435546875
2.30230236053467 -100.020500183105
2.76276278495789 -103.408599853516
3.2232232093811 -105.34984588623
3.68368363380432 -106.578010559082
4.14414405822754 -107.09147644043
4.58158159255981 -106.740219116211
4.60460472106934 -106.712821960449
5.04204225540161 -105.514892578125
5.06506490707397 -105.378623962402
5.52552556991577 -97.3811645507812
5.98598575592041 -82.7204437255859
6.44644641876221 -61.2416725158691
6.906907081604 -22.0324935913086
7.45945930480957 45.1609420776367
7.89689683914185 94.8783798217773
8.26526546478271 134.32194519043
8.28828811645508 135.848922729492
8.72572612762451 133.169448852539
8.74874877929688 132.038055419922
9.20920944213867 77.4742279052734
9.66967010498047 42.0099868774414
10.1301298141479 25.4587116241455
10.5675678253174 19.581392288208
10.5905904769897 19.5158081054688
11.0510511398315 23.9011459350586
11.5115118026733 27.2109489440918
11.9719715118408 29.4452209472656
12.4554557800293 30.6770896911621
12.9159154891968 31.6511211395264
13.3303298950195 32.3632392883301
13.3533535003662 32.4914512634277
13.7907905578613 36.1472320556641
13.8368368148804 36.7766075134277
14.2512512207031 42.9912338256836
14.2742738723755 43.8286285400391
14.7117118835449 65.4701309204102
14.757758140564 69.0499725341797
15.2182178497314 107.198127746582
15.6556558609009 137.901489257812
16.0930938720703 163.001113891602
16.1161155700684 163.369003295898
16.5535526275635 162.597808837891
16.5765762329102 161.576858520508
17.2442436218262 121.308631896973
17.6816825866699 95.7949066162109
18.1421413421631 69.7332916259766
18.4644641876221 52.1377601623535
18.9249248504639 29.2084808349609
19.362361907959 9.56418228149414
19.8228225708008 -8.68426704406738
20.2832832336426 -24.4933090209961
20.7207202911377 -37.3236122131348
21.1811809539795 -48.4314765930176
21.6416416168213 -57.4496726989746
22.1021022796631 -64.4142456054688
22.5625629425049 -70.0410766601562
23 -74.1791915893555
};
\addlegendentry{K=50}
\addplot [semithick, darkorange25512714, dashed]
table {%
0 -89.7076110839844
0.943943977355957 -102.404815673828
1.97997999191284 -112.664199829102
2.76276278495789 -120.120399475098
3.68368363380432 -123.145240783691
4.58158159255981 -123.534996032715
4.60460472106934 -123.51879119873
5.50250244140625 -118.79963684082
5.52552556991577 -118.507827758789
6.42342329025269 -86.0327835083008
6.44644641876221 -84.83203125
7.39039039611816 3.69972920417786
8.26526546478271 99.6076049804688
8.28828811645508 101.115272521973
9.18618583679199 89.4567947387695
9.20920944213867 88.7125854492188
10.1071071624756 33.6399116516113
10.1301298141479 32.8881683349609
11.0510511398315 36.4285697937012
11.9719715118408 39.2134017944336
12.8928928375244 41.2662658691406
13.7907905578613 44.1628112792969
13.813814163208 44.5594520568848
14.7117118835449 68.4099655151367
14.7347345352173 70.3300552368164
15.6326322555542 173.919738769531
15.6556558609009 174.823348999023
16.5535526275635 177.900253295898
16.5765762329102 176.900024414062
17.5665664672852 116.187698364258
18.4184188842773 67.2121658325195
19.3393402099609 23.6039810180664
20.2602596282959 -10.096302986145
21.1581573486328 -34.0074882507324
21.3193187713623 -36.7692565917969
22.0790786743164 -49.7056350708008
22.3783779144287 -52.8796615600586
23 -59.459545135498
};
\addlegendentry{K=25}
\end{axis}

\end{tikzpicture}}
  \caption{Bike-Sharing Dataset. Feature effect plots on \(X_{\texttt{hour}}\): (Left) DALE vs ALE for \(K=200\). (Center) DALE plots for \(K = \{25, 50, 100\}\). (Right) ALE plots for \(K = \{25, 50, 100\}\)}
  \label{fig:bike-sharing-feature-3}
\end{figure}

An exception is feature \(X_{\mathtt{hour}}\). In this case, the \(\dale\) approximation remains accurate when lowering the number of bins \(K\) (Fig.~\ref{fig:bike-sharing-feature-3}(b)), while \(\hat{f}_{\mathtt{ALE}}\) deteriorates significantly (Fig.~\ref{fig:bike-sharing-feature-3}(c)). In Table~\ref{tab:bike-sharing-accuracy} we evaluate both approximations on \(X_{\mathtt{hour}}\), for different number of bins \(K\). We set the ground-truth effect to be the approximation for \(K=200\). We observe that NMSE remains low in DALE for all \(K\), while for ALE it rapidly increases. This is due to OOD sampling that occurs when the bin size becomes large.

\begin{table}
  \caption{Evaluation of DALE and ALE approximation when lowering the number of bins \(K\). The ground-truth effect has been computed for \(K=200\).}
  \label{tab:bike-sharing-accuracy} \centering
  \begin{tabular}{c|c|c|c|c|c}
    \multicolumn{6}{c}{Accuracy on Bike-Sharing Dataset - Feature \(X_{\mathtt{hour}}\)} \\
    \hline \hline & & \multicolumn{4}{|c}{Number of bins} \\
    \hline & & 100 & 50 & 25 & 15 \\
    \hline \hline \multirow{2}{*}{\(\mathtt{NMSE}\)} & \(\alep\) & 0.04 & 0.43 & 0.79 & 0.83 \\
                  & \(\dale\) & \textbf{0.007} & \textbf{0.01} & \textbf{0.03} & \textbf{0.09} \\
    \hline
  \end{tabular}
\end{table}

\section{Conclusion and Future Work} This paper introduced DALE, an efficient and robust to OOD approximation for ALE. Although ALE models the feature effect correctly in cases of correlated features, ALE's approximation scales poorly in high-dimensional datasets and suffers from OOD sampling. As discussed in the paper, DALE addresses these limitations providing a fast and on-distribution alternative. We proved that under some hypotheses, our proposal is an unbiased estimator of ALE and we presented a method for quantifying the standard error of the approximation. The experiments verified our claims. DALE significantly improves the efficiency of ALE's approximation by orders of magnitude and secures that local effect estimations come from on-distribution samples. The latter leads to more accurate feature effect plots when the bins are wide and the black-box function changes away from the data generating distribution.

The computational efficiency of DALE delivers a substantial margin for future extensions. A significant advantage of our proposal is that effects are computed once on the training set points and can be reused in different-size bins. The decision for the bin density, i.e., the resolution of the plot, can be taken afterwards. Therefore, DALE permits creating feature effect plots at different resolutions with near-zero computational overhead, which can be embedded into a multi-resolution feature effect plots framework.

\acks{We thank Eirini Ntoutsi for her insightful comments in the draft versions of the paper and Giorgos Giannopoulos for the valuable discussions while forming the initial idea.

  \noindent This work was partially supported by projects i4metal (grant T2EDK- 03010) and XMANAI (grant agreement No 957362), which have received funding by the European Regional Development Fund of the EU and Greek national funds (through the Operational Program Competitiveness, Entrepreneurship and Innovation, under the call Research-Create-Innovate) and the EU 2020 Programme, ICT-38-2020 - Artificial intelligence for manufacturing, respectively.  }

\bibliography{gkolemis22-bib}

\begin{thebibliography}{19}
\providecommand{\natexlab}[1]{#1}
\providecommand{\url}[1]{\texttt{#1}}
\expandafter\ifx\csname urlstyle\endcsname\relax
  \providecommand{\doi}[1]{doi: #1}\else
  \providecommand{\doi}{doi: \begingroup \urlstyle{rm}\Url}\fi

\bibitem[Adadi and Berrada(2018)]{Adadi2018}
Amina Adadi and Mohammed Berrada.
\newblock Peeking inside the black-box: a survey on explainable artificial
  intelligence (xai).
\newblock \emph{IEEE access}, 6:\penalty0 52138--52160, 2018.

\bibitem[Apley and Zhu(2020)]{Apley2020}
Daniel~W Apley and Jingyu Zhu.
\newblock Visualizing the effects of predictor variables in black box
  supervised learning models.
\newblock \emph{Journal of the Royal Statistical Society: Series B (Statistical
  Methodology)}, 82\penalty0 (4):\penalty0 1059--1086, 2020.

\bibitem[Arrieta et~al.(2020)Arrieta, D{\'\i}az-Rodr{\'\i}guez, Del~Ser,
  Bennetot, Tabik, Barbado, Garc{\'\i}a, Gil-L{\'o}pez, Molina, Benjamins,
  et~al.]{BarredoArrieta2020}
Alejandro~Barredo Arrieta, Natalia D{\'\i}az-Rodr{\'\i}guez, Javier Del~Ser,
  Adrien Bennetot, Siham Tabik, Alberto Barbado, Salvador Garc{\'\i}a, Sergio
  Gil-L{\'o}pez, Daniel Molina, Richard Benjamins, et~al.
\newblock Explainable artificial intelligence (xai): Concepts, taxonomies,
  opportunities and challenges toward responsible ai.
\newblock \emph{Information fusion}, 58:\penalty0 82--115, 2020.

\bibitem[Baniecki et~al.(2021)Baniecki, Kretowicz, and Biecek]{Baniecki2022}
Hubert Baniecki, Wojciech Kretowicz, and Przemyslaw Biecek.
\newblock Fooling partial dependence via data poisoning.
\newblock \emph{arXiv preprint arXiv:2105.12837}, 2021.

\bibitem[Fanaee-T and Gama(2013)]{BikeSharing}
Hadi Fanaee-T and Joao Gama.
\newblock Event labeling combining ensemble detectors and background knowledge.
\newblock \emph{Progress in Artificial Intelligence}, pages 1--15, 2013.
\newblock ISSN 2192-6352.

\bibitem[Fisher et~al.(2019)Fisher, Rudin, and Dominici]{Fisher2019}
Aaron Fisher, Cynthia Rudin, and Francesca Dominici.
\newblock All models are wrong, but many are useful: Learning a variable's
  importance by studying an entire class of prediction models simultaneously.
\newblock \emph{J. Mach. Learn. Res.}, 20\penalty0 (177):\penalty0 1--81, 2019.

\bibitem[Friedman(2001)]{Friedman2001}
Jerome~H Friedman.
\newblock Greedy function approximation: a gradient boosting machine.
\newblock \emph{Annals of statistics}, pages 1189--1232, 2001.

\bibitem[Friedman and Popescu(2008)]{Friedman2008}
Jerome~H Friedman and Bogdan~E Popescu.
\newblock Predictive learning via rule ensembles.
\newblock \emph{The annals of applied statistics}, pages 916--954, 2008.

\bibitem[Gurumoorthy et~al.(2019)Gurumoorthy, Dhurandhar, Cecchi, and
  Aggarwal]{Gurumoorthy2019}
Karthik~S Gurumoorthy, Amit Dhurandhar, Guillermo Cecchi, and Charu Aggarwal.
\newblock Efficient data representation by selecting prototypes with importance
  weights.
\newblock In \emph{2019 IEEE International Conference on Data Mining (ICDM)},
  pages 260--269. IEEE, 2019.

\bibitem[Hoffman et~al.(2018)Hoffman, Mueller, Klein, and Litman]{Hoffman2018}
Robert~R Hoffman, Shane~T Mueller, Gary Klein, and Jordan Litman.
\newblock Metrics for explainable ai: Challenges and prospects.
\newblock \emph{arXiv preprint arXiv:1812.04608}, 2018.

\bibitem[Hooker et~al.(2021)Hooker, Mentch, and Zhou]{Hooker2021}
Giles Hooker, Lucas Mentch, and Siyu Zhou.
\newblock Unrestricted permutation forces extrapolation: variable importance
  requires at least one more model, or there is no free variable importance.
\newblock \emph{Statistics and Computing}, 31\penalty0 (6):\penalty0 1--16,
  2021.

\bibitem[Kim et~al.(2016)Kim, Khanna, and Koyejo]{Kim2016}
Been Kim, Rajiv Khanna, and Oluwasanmi~O Koyejo.
\newblock Examples are not enough, learn to criticize! criticism for
  interpretability.
\newblock \emph{Advances in neural information processing systems}, 29, 2016.

\bibitem[Lundberg and Lee(2017)]{Lundberg2017}
Scott~M Lundberg and Su-In Lee.
\newblock A unified approach to interpreting model predictions.
\newblock \emph{Advances in neural information processing systems}, 30, 2017.

\bibitem[Molnar et~al.(2019)Molnar, Casalicchio, and Bischl]{Molnar2021}
Christoph Molnar, Giuseppe Casalicchio, and Bernd Bischl.
\newblock Quantifying model complexity via functional decomposition for better
  post-hoc interpretability.
\newblock In \emph{Joint European Conference on Machine Learning and Knowledge
  Discovery in Databases}, pages 193--204. Springer, 2019.

\bibitem[Molnar et~al.(2020)Molnar, Casalicchio, and Bischl]{Molnar2020}
Christoph Molnar, Giuseppe Casalicchio, and Bernd Bischl.
\newblock Interpretable machine learning--a brief history, state-of-the-art and
  challenges.
\newblock In \emph{Joint European Conference on Machine Learning and Knowledge
  Discovery in Databases}, pages 417--431. Springer, 2020.

\bibitem[Nanfack et~al.(2021)Nanfack, Temple, and
  Fr{\'e}nay]{nanfack2021global}
G{\'e}raldin Nanfack, Paul Temple, and Beno{\^\i}t Fr{\'e}nay.
\newblock Global explanations with decision rules: a co-learning approach.
\newblock In \emph{Uncertainty in Artificial Intelligence}, pages 589--599.
  PMLR, 2021.

\bibitem[Ribeiro et~al.(2016)Ribeiro, Singh, and Guestrin]{Ribeiro2016}
Marco~Tulio Ribeiro, Sameer Singh, and Carlos Guestrin.
\newblock " why should i trust you?" explaining the predictions of any
  classifier.
\newblock In \emph{Proceedings of the 22nd ACM SIGKDD international conference
  on knowledge discovery and data mining}, pages 1135--1144, 2016.

\bibitem[Ribeiro et~al.(2018)Ribeiro, Singh, and Guestrin]{Ribeiro2018}
Marco~Tulio Ribeiro, Sameer Singh, and Carlos Guestrin.
\newblock Anchors: High-precision model-agnostic explanations.
\newblock In \emph{Proceedings of the AAAI conference on artificial
  intelligence}, volume~32, 2018.

\bibitem[Wachter et~al.(2017)Wachter, Mittelstadt, and Russell]{Wachter2017}
Sandra Wachter, Brent Mittelstadt, and Chris Russell.
\newblock Counterfactual explanations without opening the black box: Automated
  decisions and the gdpr.
\newblock \emph{Harv. JL \& Tech.}, 31:\penalty0 841, 2017.

\end{thebibliography}

\appendix

\section{Notation List}
\label{sec:not-list}
\begin{itemize}
\item \( s \), index of the feature of interest
\item \( \mathcal{X}_s \), feature of interest as a r.v.
\item \( \mathcal{X}_c = (\mathcal{X}_{/s}, )\), the rest of the features in as a r.v.
\item \( \mathcal{X} = (\mathcal{X}_s, \mathcal{X}_c) = (\mathcal{X}_1, \cdots, \mathcal{X}_s, \cdots, \mathcal{X}_D) \), all input features as r.v.
\item \( x_s \), feature of interest
\item \( \xc \), the rest of the features
\item \( \xb = (x_s, \xc) = (x_1, \cdots, x_s, \cdots, x_D)\), all the input features
\item \( \mathbf{X} \), design matrix/training set
\item \( f(\cdot) : \R^D \rightarrow \R \), black box function
\item \( f_s(\xb) = \frac{\partial f(x_s, \xc)}{\partial x_s} \), the partial derivative of the \( s \)-th feature
\item \( D \), dimensionality of the input
\item \( N \), number of training examples
\item \( \xb^i \), \(i\)-th training example
\item \( x^i_s \), \(s\)-th feature of the i-th training example
\item \( \xci \), the rest of the features of the i-th training example
\item \( f_{\mathtt{ALE}}(x_s) : \R \rightarrow \R\), ALE definition for the \(s\)-th feature
\item \( \hat{f}_{\mathtt{DALE}}(x_s) : \R \rightarrow \R\), DALE approximation for the \(s\)-th feature
\item \( \hat{f}_{\mathtt{ALE}}(x_s) : \R \rightarrow \R\), ALE approximation for the \(s\)-th feature
\item \( z_{k-1}, z_k\), the left and right limit of the \( k\)-th bin
\item \( \mathcal{S}_k = \{ \xb^i : x^i_s \in [z_{k-1}, z_k) \}\), the set of training points that belong to the \( k\)-th bin
\item \( k_x \) the index of the bin that \( x \) belongs to
\item \( \hat{\mu}_k^s\), \(\mathtt{DALE}\) approximation of the mean value inside a bin, equals \( \frac{1}{|\mathcal{S}_k|} \sum_{i: x^i\in \mathcal{S}_k} f_s(\xb^i) \)
\item \( (\hat{\sigma}_k^s)^2\), \(\mathtt{DALE}\) approximation of the variance inside a bin, equals \( \frac{1}{|\mathcal{S}_k|-1} \sum_{i: x^i\in \mathcal{S}_k} (f_s(\xb^i) - \hat{\mu}_k^s)^2 \)

\end{itemize}

\section{Derivation of equations in the Background section}

In this section, we present the derivations for obtaining the feature
effect at the Background.

\subsubsection*{Example Definition.}The black-box function and the
generating distribution are:

\begin{equation}
  \label{eq:black-box}
  f(x_1, x_2) =
  \begin{cases}
    1 - x_1 - x_2 \: \: \:  ,\text{if} \: x_1 + x_2  \leq 1 \\
    0 \quad \quad \quad \quad \quad \:, \text{otherwise}
  \end{cases}
\end{equation}

\begin{equation}
  \label{eq:generative}
  p(\mathcal{X}_1 = x_1, \mathcal{X}_2=x_2) =
  \begin{cases}
    1 & x_1 \in [0,1], x_2=x_1 \\
    0 & \text{otherwise}
  \end{cases}
\end{equation}

\begin{equation}
  \label{eq:marginal}
  p(\mathcal{X}_1 = x_1) =
  \begin{cases}
    1 & 0 \leq x_1 \leq 1 \\
    0 & \text{otherwise}
  \end{cases}
\end{equation}

\begin{equation}
  \label{eq:marginal}
  p(\mathcal{X}_2 = x_2) =
  \begin{cases}
    1 & 0 \leq x_2 \leq 1 \\
    0 & \text{otherwise}
  \end{cases}
\end{equation}

\begin{equation}
  \label{eq:marginal}
  p(\mathcal{X}_2 = x_2|\mathcal{X}_1 = x_1) = \delta(x_2-x_1)
\end{equation}

\subsubsection*{PDPlots.}

The feature effect computed by PDP plots is:

\begin{equation}
  \label{eq:example-1-pdp}
  \begin{split}
    f_{\mathtt{PDP}}(x_1) &= \\
    & = \mathbb{\E}_{\mathcal{X}_2}[f(x_1,\mathcal{X}_2)] \\
    & = \int_{x_2} f(x_1, x_2) p(x_2) \partial x_2 \\
    & = \int_{0}^{1-x_1} (1 - x_1 - x_2) \partial x_2 + \int_{1-x_1}^1 0 \partial x_2 \\
    & = \int_{0}^{1-x_1} 1 \partial x_2 + \int_{0}^{1-x_1} -x_1 \partial x_2 + \int_{0}^{1-x_1} -x_2 \partial x_2 \\
    & = (1 - x_1) -x_1(1-x_1) - \frac{{(1-x_1)}^2}{2} \\
    & = (1 - x_1)^2 - \frac{{(1-x_1)}^2}{2} \\
    & = \frac{{(1-x_1)}^2}{2}
  \end{split}
\end{equation}

Due to symmetry:

\begin{equation}
y = f_{\mathtt{PDP}}(x_2) = \frac{{(1-x_2)}^2}{2}
\end{equation}

\subsubsection*{MPlots.}

The feature effect computed by PDP plots is:

\begin{equation}
  \label{eq:example-1-MPlots}
  \begin{split}
    f_{\mathtt{MP}}(x_1) &= \\
    & = \mathbb{\E}_{\mathcal{X}_2|\mathcal{X}_1=x_1}[f(x_1,\mathcal{X}_2)] \\
    & = \int_{x_2} f(x_1,x_2) p(x_2|x_1) \partial x_2 \\
    & =   f(x_1, x_1) = \\
  & = \begin{cases}
    1 - 2x_1, & x_1 \leq 0.5 \\
    0, & \text{otherwise}
\end{cases}
  \end{split}
\end{equation}
Due to symmetry:

\begin{equation}
  y = f_{\mathtt{MP}}(x_2) =
  \begin{cases}
    1 - 2x_2 & x_2 \leq 0.5 \\
    0, &\text{otherwise}
  \end{cases}
\end{equation}

\subsubsection*{ALE}

The feature effect computed by ALE is:

\begin{equation}
  \label{eq:example-1-ale}
  \begin{split}
    f_{\mathtt{ALE}}(x_1) &= \\
    & = \int_{z_0}^{x_1} \mathbb{E}_{\mathcal{X}_2|\mathcal{X}_1=z} \left [\frac{\partial f}{\partial z}(z, \mathcal{X}_2) \right ] \partial z \\
    & = \int_{z_0}^{x_1} \int_{x_2} \frac{\partial f}{\partial z}(z,x_2) p(x_2|z)  \partial x_2 \partial z = \\
    & = \int_{z_0}^{x_1} \frac{\partial f}{\partial z}(z,z) \partial z = \\
    & = \begin{cases}
      \int_{z_0}^{x_1} -1 \partial z & x_1 \leq 0.5 \\
      \int_{z_0}^{0.5} -1 \partial z + \int_{.5}^{x_1} 0 \partial z & x_1 > 0.5
    \end{cases} \\
    & = \begin{cases}
      -x_1 & x_1 \leq 0.5 \\
      -0.5 & x_1 > 0.5
    \end{cases}
  \end{split}
\end{equation}

The normalization constant is:

\begin{equation}
  \label{eq:constant}
  \begin{split}
    c & = - \mathbb{E}[\hat{f}_{ALE}(x_1)] \\
    & = - \int_{-\infty}^{\infty} \hat{f}_{ALE}(x_1) \\
    & = - \int_{0}^{0.5} - z \partial z - \int_{0.5}^{1} -0.5 \partial z \\
    & = \frac{0.25}{2} + 0.25 = 0.375
  \end{split}
\end{equation}

Therefore, the normalized feature effect is:

\begin{gather}
y = f_{\mathtt{ALE}}(x_1) =
\begin{cases}
0.375 - x_1 & 0 \leq x_1 \leq 0.5\\
- 0.125 &  0.5 < x_1 \leq 1
\end{cases}
\end{gather}

Due to symmetry:

\begin{gather}
y = f_{ALE}(x_2) =
\begin{cases}
0.375 - x_2 & 0 \leq x_2 \leq 0.5\\
- 0.125 &  0.5 < x_2 \leq 1
\end{cases}
\end{gather}

\section{First-order and Second-order DALE approximation}

In the main part of the paper, we presented the first order ALE approximation as

\begin{align}
  f_{\mathtt{DALE}}(x_s) = \Delta x \sum_{k=1}^{k_x} \frac{1}{|\mathcal{S}_k|}
  \sum_{i:\xb^i \in \mathcal{S}_k} [f_s(\xb^i)]
\end{align}
For keeping the equation compact, we ommit a small detail about the
manipulation of the last bin. In reality, we take complete
\( \Delta x \) steps until the \( k_x - 1 \) bin, i.e. the one that
prepends the bin where \( x \) lies in. In the last bin, instead of a
complete \( \Delta x \) step, we move only until the position \( x
\). Therefore, the exact first-order DALE approximation is

\begin{equation}
  \begin{split}
  f_{\mathtt{DALE}}(x_s) &= \Delta x \sum_{k=1}^{k_x-1} \frac{1}{|\mathcal{S}_k|}
  \sum_{i:\xb^i \in \mathcal{S}_k} [f_s(\xb^i)]\\
  & + (x-z_{(k_x-1)}) \frac{1}{|\mathcal{S}_{k_x}|} \sum_{i:\xb^i \in
    \mathcal{S}_{k_x}} [f_s(\xb^i)]
  \end{split}
  \label{eq:DALE_first_order_complete}
\end{equation}

\noindent
Following a similar line of thought we define the complete second-order DALE approximation as

\begin{multline}
  f_{\mathtt{DALE}}(x_l, x_m) = \Delta x_l\sum_{p=1}^{p_x-1} \Delta x_m\sum_{q=1}^{q_x-1} \frac{1}{|\mathcal{S}_{k,q}|} \sum_{i:\xb^i \in \mathcal{S}_{k,q}}f_{l,m}(\xb^i)\\
  + (x_l-z_{(p_x-1)})(x_m-z_{(q_x-1)}) \frac{1}{|\mathcal{S}_{p_x,q_x}|} \sum_{i:\xb^i \in \mathcal{S}_{p_x,q_x}}f_{l,m}(\xb^i)
 \label{eq:DALE_second_order_complete}
\end{multline}

\section{Second-order ALE definition}

The second-order ALE plot definintion is

\begin{equation}
  f_{\mathtt{ALE}}(x_l, x_m) = c + \int_{x_{l, min}}^{x_l} \int_{x_{m, min}}^{x_m} \mathbb{E}_{\Xc|X_l=z_l,
      X_m=z_m}[f_{l,m}(\xb)] \partial z_l \partial z_m
  \label{eq:ALE2}
\end{equation}

\noindent

where
\( f_{l,m}(\xb) = \dfrac{\partial^2f(x)}{\partial x_l \partial x_m} \).

\section{DALE variance inside each bin}

In this section, we show that the variance of the local effect
estimation inside a bin, i.e. \(\mathrm{Var}[\hat{\mu}_k^s]\) equals
with \(\frac{(\sigma_k^s)^2}{|\mathcal{S}_k|}\), where
\((\sigma_k^s)^2 = \mathrm{Var}[f_s(\mathbf{x})]\).

\begin{equation}
  \begin{split}
  \mathrm{Var}[\hat{\mu}_k^s] &= \mathrm{Var} [\frac{1}{|\mathcal{S}_k|} \sum_{i: x^i\in \mathcal{S}_k} f_s(\xb^i)] \\
                              &= \frac{1}{|\mathcal{S}_k|^2} \sum_{i: x^i\in \mathcal{S}_k} \mathrm{Var}[f_s(\xb^i)] \\
                              &= \frac{|\mathcal{S}_k|}{|\mathcal{S}_k|^2} \mathrm{Var}[f_s(\xb)] \\
  &= \frac{(\sigma_k^s)^2}{|\mathcal{S}_k|}  \\
  \end{split}
\end{equation}

\end{document}